\documentclass[final,twocolumn,times]{elsarticle}

\usepackage[T1]{fontenc}

\usepackage[hang]{subfigure}

\usepackage{xtab,booktabs}

\usepackage{tikz-cd}

\usepackage{lineno,hyperref}
\modulolinenumbers[5]

\journal{}

\usepackage{changepage}

\usepackage{amsthm} 

\usepackage{caption}

\usepackage{paralist}

\usepackage{amsmath}
\usepackage{amssymb}
\usepackage{amsfonts}
\usepackage{mathrsfs}
\usepackage{subfigure}
\usepackage{epsfig}
\usepackage{graphicx}
\usepackage{color}

\usepackage{float}

\DeclareCaptionFormat{cont}{#1 (cont.)#2#3\par}

\usepackage[linesnumbered,ruled,]{algorithm2e}

\begin{document}
	
	\begin{frontmatter}
		

		\title{DC-Reg: Globally Optimal Point Cloud Registration via Tight Bounding with Difference of Convex Programming}

		\author[mymainaddress]{Wei Lian \corref{cor1}} 
	\ead{lianwei3@gmail.com}

	\author[mymainaddress]{Fei Ma}
	\ead{mafei@czc.edu.cn}

	\author[mymainaddress]{Hang Pan}
	\ead{panhang@czc.edu.cn}
	
	\author[mymainaddress]{Zhesen Cui}
	\ead{cuizhesen@gmail.com}

	\author[mysecondaryaddress]{Wangmeng Zuo}
	\ead{cswmzuo@gmail.com}


	\cortext[cor1]{Corresponding author}

	\address[mymainaddress]{Department of Computer Science, Changzhi University, Changzhi, Shanxi, China, 046011}
	\address[mysecondaryaddress]{School of Computer Science and Technology, Harbin Institute of Technology, Harbin 150001, China}
	
	\begin{abstract}

Achieving globally optimal point cloud registration under partial overlaps and large misalignments remains a fundamental challenge. While simultaneous transformation ($\boldsymbol{\theta}$) and correspondence ($\mathbf{P}$) estimation has the advantage of being robust to nonrigid deformation, its non-convex coupled objective often leads to local minima for heuristic methods and prohibitive convergence times for existing global solvers due to loose lower bounds. To address this, we propose DC-Reg, a robust globally optimal framework that significantly tightens the Branch-and-Bound (BnB) search. Our core innovation is the derivation of a holistic concave underestimator for the  coupled
transformation-assignment
 objective, grounded in the Difference of Convex (DC) programming paradigm.
Unlike prior works that rely on term-wise relaxations (e.g., McCormick envelopes) which neglect variable interplay, our holistic DC decomposition captures the joint structural interaction between $\boldsymbol{\theta}$ and $\mathbf{P}$. This formulation enables the computation of remarkably tight lower bounds via efficient Linear Assignment Problems (LAP) evaluated at the vertices of the search boxes. We validate our framework on 2D similarity and 3D rigid registration, utilizing rotation-invariant features for the latter to achieve high efficiency without sacrificing optimality. Experimental results on synthetic data and the 3DMatch benchmark demonstrate that DC-Reg achieves significantly faster convergence and superior robustness to extreme noise and outliers compared to state-of-the-art global techniques.

	\end{abstract}
	
	
\end{frontmatter}



\section{Introduction}

Point cloud registration is a fundamental task in computer vision, yet achieving global optimality under significant noise, outliers, and non-rigid deformations remains challenging. While local methods like ICP \cite{ICP} and CPD \cite{CPD_match} are efficient, their reliance on initialization often leads to suboptimal local minima.

To ensure global convergence, Branch-and-Bound (BnB) frameworks \cite{scholz2011deterministic} have been widely adopted. However, correspondence-free solvers like Go-ICP \cite{Go-ICP_pami} are largely restricted to rigid alignment. Better  paradigm involves simultaneous estimation of pose ($\boldsymbol{\theta}$) and correspondences ($\mathbf{P}$) \cite{RPM_TPS}, but its coupled non-convex objective is notoriously difficult to optimize. Current global solvers face a critical trade-off 
between two strategies:

\begin{inparaenum}[\upshape i\upshape)]
	
	\item \textbf{Variable Elimination:} By eliminating the transformation parameters $\boldsymbol{\theta}$, this approach reduces the problem to a concave minimization over the correspondence matrix $\mathbf{P}$.
	 While mathematically elegant, these methods are often restricted to subset-embedding scenarios \cite{RPM_concave, RPM_concave_PAMI} or suffer from high-dimensional branching within the projected space of the correspondence variables, which scales poorly as problem size  increases
	\cite{RPM_model_occlude, RPM_model_occlude_PR}.


\item \textbf{Term-wise Relaxation:} These methods perform branching in the low-dimensional transformation space while retaining both $\boldsymbol{\theta}$ and $\mathbf{P}$ within the optimization. However, they typically derive lower bounds by relaxing each objective term independently---for instance, via convex envelopes of trilinear \cite{LIAN2023126482} or bilinear monomials \cite{Lian2024}. This "term-wise" independence neglects the global structural interplay between variables, leading to loose lower bounds and prohibitive convergence times, especially in outlier-heavy regimes.

\end{inparaenum}

In this paper, we follow the second strategy by branching exclusively in the transformation space. To overcome the limitations of independent relaxations, we propose DC-Reg, a globally optimal framework that significantly tightens the BnB search. By leveraging the Difference of Convex (DC) programming paradigm, we derive a holistic concave underestimator for the entire coupled objective.

Our primary contributions are:

\begin{inparaenum}[\upshape i\upshape)]
	\item \textbf{Holistic DC Bounding:} Instead of term-wise relaxation, we linearize the convex component of a holistic DC decomposition. This generates demonstrably tighter bounds that capture the joint interaction of variables, significantly accelerating the pruning process.
	
	\item \textbf{Algorithmic Flexibility:} Our framework seamlessly handles 2D similarity and 3D registration (via rotation-invariant features \cite{10.1007/978-3-030-01258-8_28}), maintaining high resilience to non-rigid distortions where rigid-only solvers fail.
	
	\item \textbf{Superior Robustness:} DC-Reg ensures reliable performance even at  low inlier ratios, outperforming state-of-the-art global solvers on both synthetic benchmarks and the 3DMatch dataset.
\end{inparaenum}

\section{Related Work \label{sec:relate_work}}

Point set registration is a mature field with diverse optimization strategies; for a comprehensive survey, see~\cite{10.1007/s11263-020-01359-2}. We categorize the most relevant works into local search, global optimization, and robust pre-filtering.

\textbf{Local and Robust Optimization.} Local methods such as ICP \cite{besl1992method} and its variants \cite{rusinkiewicz2001efficient} are the standards for fine alignment but remain sensitive to initialization. To enhance robustness, Iteratively Reweighted Least-Squares (IRLS) and M-estimators have been rigorously analyzed regarding convergence on manifolds and global rates in outlier-robust estimation \cite{aftab2015convergence, peng2023block, peng2022global}. Alternatively, Graduated Non-Convexity (GNC) avoids local minima by solving increasingly non-convex surrogates, as seen in Fast Global Registration (FGR) \cite{zhou2016fast} and robust spatial perception frameworks \cite{yang2020graduated}. While efficient, these methods lack the mathematical guarantees of global optimality required for extreme misalignments.

\textbf{BnB-based Global Methods.} Branch-and-Bound (BnB) \cite{scholz2011deterministic} provides a deterministic path to the global optimum. Early applications focused on 3D-3D registration \cite{li20073d} and Euclidean problems \cite{olsson2008branch}, eventually leading to Go-ICP \cite{yang2016go}. Recent innovations have accelerated BnB by leveraging rotation-invariant features \cite{liu2018efficient}, stereographic projections \cite{parra2016fast}, and novel transformation decompositions \cite{chen2022deterministic, wang2022efficient}. However, many existing BnB solvers either decouple the problem into sequential searches or employ term-wise relaxations \cite{LIAN2023126482}, which often yield loose bounds and slow convergence in the presence of heavy outliers.

\textbf{Certifiable and Convex Relaxations.} Semidefinite Programming (SDP) relaxations offer a "certifiably optimal" alternative by transforming non-convex registration into convex programs \cite{briales2017convex, iglesias2020global}. Notably, Yang and Carlone \cite{yang2019quaternion} achieved resilience against extreme outlier ratios through quaternion-based Wahba solvers. While theoretically elegant, the tightness of these relaxations can be sensitive to noise models \cite{peng2022semidefinite}, and their computational cost often exceeds that of modern geometric solvers.

\textbf{Outlier Removal and Consistency Graphs.} To manage computational complexity, pre-processing steps like Guaranteed Outlier Removal (GORE) \cite{parra2018guaranteed} and Mixed-Integer Linear Programs (MILP) \cite{chin2016guaranteed} are used to prune the search space. Modern robust solvers frequently leverage consistency graphs to identify reliable correspondences. TEASER \cite{yang2021teaser} and CLIPPER \cite{lusk2021clipper} utilize truncated least-squares and graph-theoretic data association, respectively. Further extensions like SC2-PCR++ \cite{chen2023sc2pcrplus} and MAC \cite{zhang20233d} exploit high-order spatial compatibility and maximal cliques to find the largest consensus set of correspondences.

\textbf{Deep Learning for Registration.} Data-driven approaches have evolved from direct regression models like DCP \cite{wang2019deep} to frameworks that integrate robust geometric encoding. Deep Global Registration (DGR) \cite{choy2020deep} employs 6D convolutional networks with weighted solvers, while recent works like Hunter \cite{yao2023hunter} and Qin et al. \cite{qin2023deep} utilize Transformers and graph-based consistency to handle extreme outlier ratios and non-rigid distortions.

\textbf{Our Approach.} Unlike local M-estimators or decoupled GNC solvers, DC-Reg provides a deterministic global guarantee via BnB. Distinguishing ourselves from previous BnB methods that rely on loose, term-wise relaxations \cite{LIAN2023126482,Lian2024}, we utilize a holistic DC programming paradigm. This allows us to capture the joint interaction between pose and correspondence, providing a tighter lower bound that significantly accelerates convergence compared to existing global registration techniques.

\section{Methodology: The DC-Reg Framework}

We propose DC-Reg, a globally optimal registration framework employing a Branch-and-Bound (BnB) strategy. The core innovation lies in deriving a tight lower bound for the coupled objective function via a holistic Difference of Convex (DC) decomposition.

\subsection{General Bounding Operation}
\textbf{Problem Formulation.} We aim to minimize the coupled transformation-assignment objective:
\begin{equation}
	E(\mathbf{P}, \boldsymbol{\theta}) = \sum_{i,j} p_{ij} E_{ij}(\boldsymbol{\theta})
\end{equation}
where $\boldsymbol{\theta} \in \mathbb{R}^{n_\theta}$ denotes the transformation parameters constrained within a search box $\mathcal{M} = [\underline{\boldsymbol{\theta}}, \overline{\boldsymbol{\theta}}]$, and $\mathbf{P}=\{p_{ij}\}$ is the assignment matrix belonging to the feasible set $\mathcal{P}$. The set $\mathcal{P}$ typically enforces partial permutation constraints, allowing for one-to-one matching while accommodating outliers.

Minimizing the joint objective $E(\mathbf{P}, \boldsymbol{\theta})$ over both variables is equivalent to minimizing the marginalized objective function $E(\boldsymbol{\theta})$:
\begin{equation}
	E(\boldsymbol{\theta}) = \min_{\mathbf{P} \in \mathcal{P}} \sum_{i,j} p_{ij} E_{ij}(\boldsymbol{\theta})
\end{equation}

\begin{figure}[!t]
		\centering		
		\includegraphics[trim={0.1cm .5cm 3.9cm 1.1cm}, clip,  
		 width=0.45\textwidth]{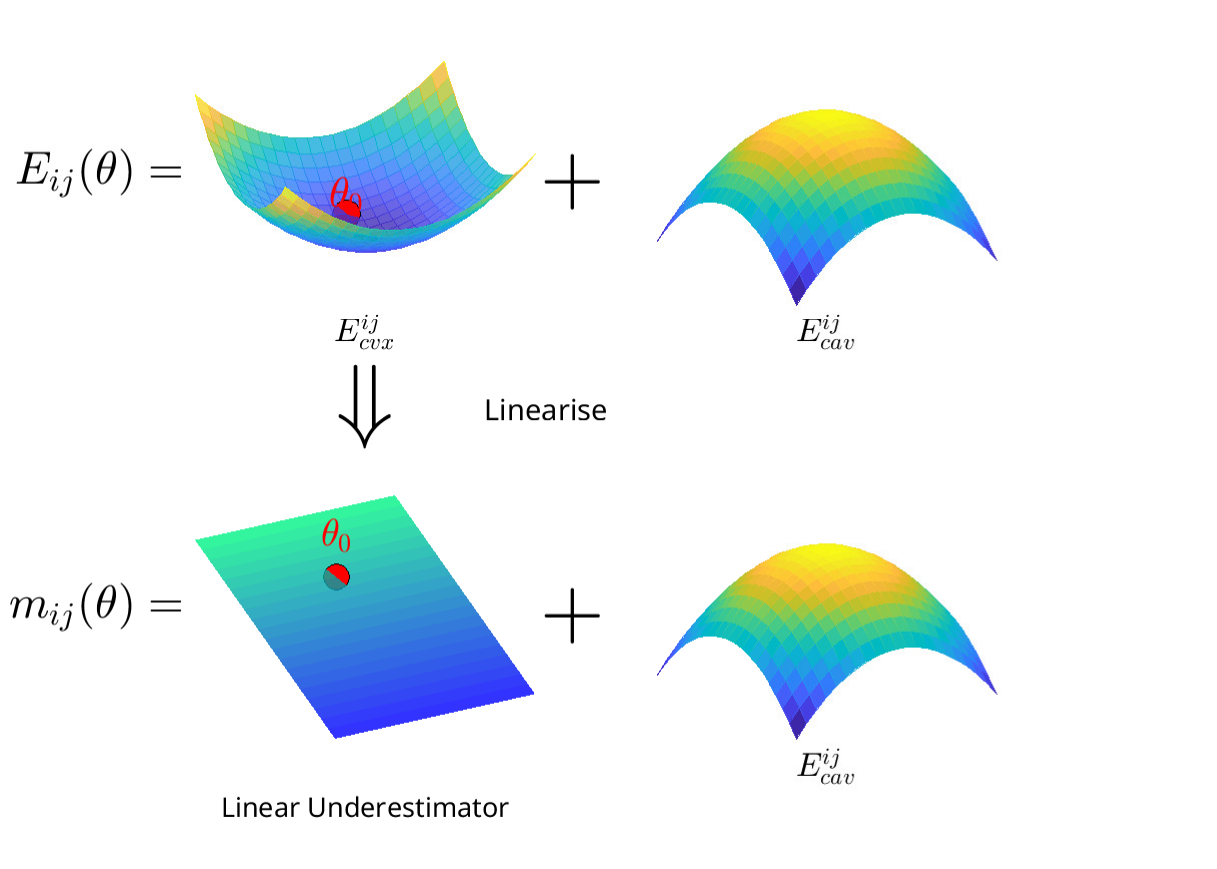}	
		\caption{
obain concave underestimator $m_{ij}$ of $E_{ij}$ via DC decomposition and linearization of the convex term.
			\label{DC_decompo}	}
\end{figure}

\textbf{DC Decomposition and Concave Underestimation.} To bound $E(\boldsymbol{\theta})$ over a sub-box $M$, we assume the pairwise cost $E_{ij}(\boldsymbol{\theta})$ admits a DC decomposition:\begin{equation}E_{ij}(\boldsymbol{\theta}) = E^{ij}_{\text{cvx}}(\boldsymbol{\theta}) + E^{ij}_{\text{cav}}(\boldsymbol{\theta})\end{equation}
where $E^{ij}_{\text{cvx}}$ is convex and $E^{ij}_{\text{cav}}$ is concave. We construct a concave underestimator $m_{ij}(\boldsymbol{\theta})$ of $E_{ij}(\theta)$ by linearizing the convex term $E^{ij}_{\text{cvx}}$ at a point  $\boldsymbol{\theta}_0\in  M$:
\begin{equation}
	m_{ij}(\boldsymbol{\theta}) := \underbrace{E^{ij}_{\text{cvx}}(\boldsymbol{\theta}_0) +
		\xi^\top
		(\boldsymbol{\theta} - \boldsymbol{\theta}_0)}_{\text{Linear Underestimator }} + E^{ij}_{\text{cav}}(\boldsymbol{\theta})
\end{equation}
where   $\xi $ is a subgradient of $E_{cvx}^{ij}(\theta)$ at $\theta_0$.
If $E_{cvx}^{ij}(\theta)$ is differentiable,
the gradient $\nabla  E_{cvx}^{ij}(\theta_0)$ is the  unique subgradient $\xi$. 
Fig. \ref{DC_decompo} illustrates the process of constructing the concave underestimator $m_{ij}(\boldsymbol{\theta})$ of $E_{ij}(\boldsymbol{\theta})$.


\begin{figure}[!t]
		\centering
		\begin{tabular}{@{\hspace{-4mm}}c} 
			\includegraphics[width=.808\linewidth]{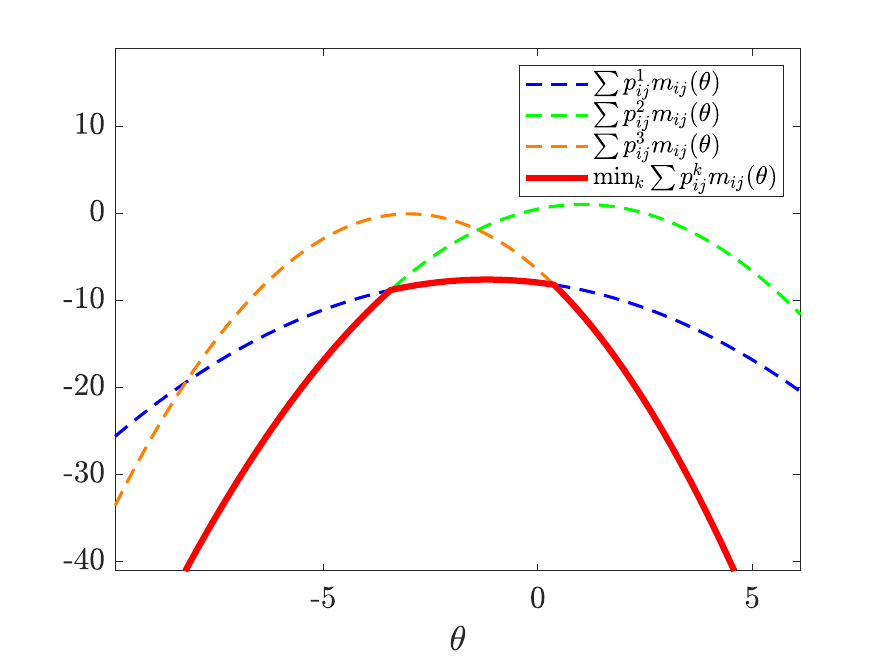}			\end{tabular}	
		\caption{
			Pointwise minimum of concave functions is a concave function.
			\label{pointwise_min_concave}	}
\end{figure}

\textbf{Lower Bound Calculation.} We define an  auxiliary function 
\begin{equation}
	z(\boldsymbol{\theta}) = \min_{\mathbf{P} \in \mathcal{P}} \sum_{i,j} p_{ij} m_{ij}(\boldsymbol{\theta})   \label{eq:z_fun}
\end{equation}
Since $m_{ij}(\boldsymbol{\theta})$ is concave,
therefore, for any fixed $\mathbf{P}\ge 0$, the term $\sum p_{ij} m_{ij}(\boldsymbol{\theta})$ is concave. 
Consequently, $z(\boldsymbol{\theta})$ is the pointwise minimum of concave functions, implying $z(\boldsymbol{\theta})$ is itself concave,
as illustrated in Fig. \ref{pointwise_min_concave}.
 A fundamental property of concave minimization over a polytope dictates that the minimum is attained at a vertex. Thus, a valid lower bound $LB(M)$ for $E(\boldsymbol{\theta})$ over $M$ is:
\begin{equation}
	LB(M) := \min_{\boldsymbol{v} \in V(M)} z(\boldsymbol{v}) \label{eq:lb_prob}
\end{equation}
where $V(M)$ is the set of the $2^{n_\theta}$ vertices of $M$. This reduces the bounding step to solving a Linear Assignment Problem (LAP) at each vertex  of the box $M$.

\subsection{Branch-and-Bound Algorithm}


Our BnB procedure (Algorithm \ref{alg:bnb}) iteratively partitions the search space. By using the tight lower bound derived in \eqref{eq:lb_prob}, we efficiently prune sub-optimal regions.

%
%
	%
	%
	%
	%
		%
		%
		%
%

\begin{algorithm}
	\caption{A BnB algorithm  for minimizing $E(\theta)$ \label{alg:bnb}}
	\textbf{Initialization:}	
	Set tolerance error $\epsilon>0$	and
	initial  box  $M_{init} = [\underline{\boldsymbol{\theta}},\overline{\boldsymbol{\theta}}]$. 	
	Let $\mathscr M_1=\mathscr N_1=\{M\}$.	
	
	\For{$k=1,2,\ldots$}{
		
		\textbf{Bounding:}
		For each  $M\in \mathscr N_k$,
		compute lower bound $LB(M)$ 
		and solution $\boldsymbol{\theta}(M)$		
		via Eq. \eqref{eq:lb_prob}.
		
		\textbf{Update Upper Bound:}
		Let $\mathbf \theta^k=\arg\min\{E(\theta^{k-1}), E(\theta(M),M\in \mathscr N_k\}$. 
		
		\textbf{Pruning:}
		Delete all $M\in\mathscr M_k$ such that $LB(M)\ge E(\mathbf \theta^k)-\epsilon$.
		Let $\mathscr R_k$ be the remaining collection of boxs.
		
		\textbf{Termination:}
		If $\mathscr R_k=\emptyset$,
		terminate: $\mathbf \theta^k $ is the global $\epsilon-$minimum solution.
		
		\textbf{Branching:}
		Let  $M^*= \arg\min\{LB (M)|M\in \mathscr R_k \}$. Divide $M^*$ into  $M_1$, $M_2$
		by bisecting  the longest edge.
		
		\textbf{Iteration:}
		Let $\mathscr N_{k+1}=\{M_1,M_2\}$
		and $\mathscr M_{k+1}=(\mathscr R_k\backslash M) \cup \mathscr N_{k+1}$.

	}

\end{algorithm}

%
%
%
	%
	%
	%
	%
	%
	%
	%
	%
	%
	%
	%
%
%

\section{Applications}
We apply the DC-Reg framework to two specific registration scenarios.

\subsection{Case I: 2D Similarity Registration}For 2D similarity registration, we parameterize the transformation by $\boldsymbol{\theta} \in \mathbb{R}^4$, where $[\theta_3, \theta_4]^\top$ is the translation vector and $\{\theta_1, \theta_2\}$ encode the scale $s$ and rotation angle $\phi$ via $\theta_1 = s\cos\phi$ and $\theta_2 = s\sin\phi$. Under this model, we adopt the coupled linear assignment-least squares objective \cite{RPM_TPS}:
\begin{equation}E(\mathbf{P}, \boldsymbol{\theta}) = \sum_{i,j} p_{ij}  \underbrace{\| \mathbf{J}_i \boldsymbol{\theta} - \mathbf{y}_j \|^2 }_{E_{ij}(\mathbf{\boldsymbol{\theta}})}
\end{equation}
where $E_{ij}(\boldsymbol{\theta})$ represents the residual error under the spatial mapping for source point $\mathbf{x}_i$ and target point $\mathbf{y}_j$. The Jacobian matrix is defined as $\mathbf J_i \triangleq \begin{bmatrix} x_i^1 & -x_i^2 & 1 & 0 \\ x_i^2 & x_i^1 & 0 & 1 \end{bmatrix}$.

Since $E_{ij}(\boldsymbol{\theta})$ is inherently convex in $\boldsymbol{\theta}$, its Difference of Convex (DC) decomposition is straightforward, with the concave component being zero ($E^{ij}_{\text{cav}} = 0$). Consequently, a concave underestimator $m_{ij}(\boldsymbol{\theta})$ can be obtained via the first-order Taylor expansion of $E_{ij}(\boldsymbol{\theta})$ at a reference point $\boldsymbol{\theta}_0 \in \mathcal{M}$:
\begin{equation}
	m_{ij}(\boldsymbol{\theta}) := E_{ij}(\boldsymbol{\theta}_0) + \nabla E_{ij}(\boldsymbol{\theta}_0)^\top (\boldsymbol{\theta} - \boldsymbol{\theta}_0)
\end{equation}
where the gradient is derived as  $ \nabla E_{ij}(\boldsymbol{\theta}_0) = 2 \mathbf{J}_i^\top (\mathbf{J}_i \boldsymbol{\theta}_0 - \mathbf{y}_j)
$.

While this formulation is highly robust to outliers and noise, extending it directly to 3D affine cases ($\boldsymbol{\theta} \in \mathbb{R}^{12}$) is computationally prohibitive. The exponential growth of the search space—requiring $2^{12}=4096$ vertex evaluations and a corresponding number of Linear Assignment Problems (LAP) per Branch-and-Bound iteration---renders the global search intractable for such high-dimensional parameters.

\subsection{Case II: Efficient 3D Registration via RIFs}

To handle 3D registration efficiently, we decouple the problem using Rotation-Invariant Features (RIFs) \cite{10.1007/978-3-030-01258-8_28}, specifically point norms, to solve for translation $\mathbf{t} \in \mathbb{R}^3$ first.

\textbf{Objective.} To achieve global optimality in 3D translation estimation independently of rotation, we minimize the mismatch between the norms of translated source points and target points. The resulting coupled translation-assignment objective is formulated as:
\begin{equation}
	E(\mathbf{P}, \mathbf{t}) = \sum_{i,j} p_{ij} \underbrace{\left( \|\mathbf{x}_{i} + \mathbf{t}\| - \|\mathbf{y}_{j}\| \right)^2}_{E_{ij}(\mathbf{t})}
\end{equation}
where
 $E_{ij}(\mathbf{t})$ represents the residual error in the RIF (Rotation-Invariant Feature) space. 


\textbf{DC Decomposition.} Expanding $E_{ij}(\mathbf{t})$ reveals a natural DC structure:
\begin{subequations}
\begin{align}
	E^{ij}_{\text{cvx}}(\mathbf{t}) &= \|\mathbf{x}_i+\mathbf{t}\|^2 \\
	E^{ij}_{\text{cav}}(\mathbf{t}) &= - 2\|\mathbf{y}_j\| \cdot \|\mathbf{x}_i+\mathbf{t}\| + \|\mathbf{y}_j\|^2
\end{align}
\end{subequations}


\textbf{Bounding.}
We linearize  $E^{ij}_{\text{cvx}}(\mathbf{t})$ at a point $\mathbf{t}_0 \in M$:
\begin{equation}
E^{ij}_{\text{cvx}}(\mathbf{t}) \ge
E^{ij}_{\text{cvx}}(\mathbf{t_0}) +  \nabla E^{ij}_{\text{cvx}}(\mathbf{t}_0) ^\top (\mathbf{t}-\mathbf{t}_0)
\end{equation}
with gradient $ \nabla E^{ij}_{\text{cvx}}(\mathbf{t}_0)=2(\mathbf{x}_i+\mathbf{t}_0).$
Combining this linear underestimator with the concave term $E^{ij}_{\text{cav}}(\mathbf{t})$ yields the concave underestimator $m_{ij}(\mathbf{t})$:
\begin{equation}
m_{ij}(\mathbf{t}) := \underbrace{
	E^{ij}_{\text{cvx}}(\mathbf{t_0}) + 2(\mathbf{x}_i+\mathbf{t}_0)^\top (\mathbf{t}-\mathbf{t}_0)}_{\text{Linear Underestimator of } E^{ij}_{\text{cvx}}} 
+ E^{ij}_{\text{cav}}(\mathbf{t})
\end{equation}

Calculating $LB(M)$ now requires checking only $2^3=8$ vertices, which is computationally highly efficient.



\textbf{Rotation Recovery.} Once the optimal translation $\mathbf{t}^*$ is recovered, the remaining rotation $\mathbf{R}$ can be estimated using any  rotation search algorithm. In this work, we adopt the method from \cite{Lian2024} for its  robustness to non-rigid deformations.

\section{Implementation Details}
\textbf{Linearization Point:} We consistently set $\boldsymbol{\theta}_0$ (or $\mathbf{t}_0$) as the geometric center of the current box $M$, which minimizes the maximum approximation error over the domain.

\textbf{Constraints:} We enforce one-to-one matching with a fixed inlier count $n_p$:
$\mathcal{P} = \{\mathbf{P} \in \{0,1\}^{N \times M} \mid \mathbf{P}\mathbf{1} \le \mathbf{1}, \mathbf{1}^\top \mathbf{P} \le \mathbf{1}^\top, \sum p_{ij} = n_p\}$.
The resulting vertex evaluation in Eq. \eqref{eq:lb_prob} is an $n_p-$cardinality Linear Assignment Problem (LAP), which we solve efficiently using the Jonker-Volgenant algorithm.

\section{Experiments and Results \label{sec:exp}}

We implement the proposed DC-Reg framework in Matlab R2023b. All experiments are conducted on a workstation equipped with a 6-core 3.2 GHz CPU.
The registration performance is evaluated using the Root Mean Square Error (RMSE) between the transformed source inliers and their corresponding target inliers.



\subsection{2D Registration Experiments} We evaluate DC-Reg against state-of-the-art global solvers: RPM-HTB \cite{LIAN2023126482}, RPM-PA \cite{lian2021polyhedral}, and RPM-CAV \cite{RPM_model_occlude_PR}. These methods share our ability to handle partial overlap and recover arbitrary similarity transformations via global optimization, making them suitable for direct comparison.

\subsubsection{2D Synthetic Data Analysis} 
\begin{figure*} [!th]
	\setlength{\abovecaptionskip}{-1pt plus 0pt minus 12pt} 
	\centering
	\newcommand{\scale}{0.10	}

	\begin{tabular}{@{\hspace{-0mm}}c@{}|@{}c@{}|@{}c@{}|@{}c@{}|@{}c@{}|@{}c@{}|@{}c@{}|@{}c|@{}c }	
		\includegraphics[width=\scale\linewidth]{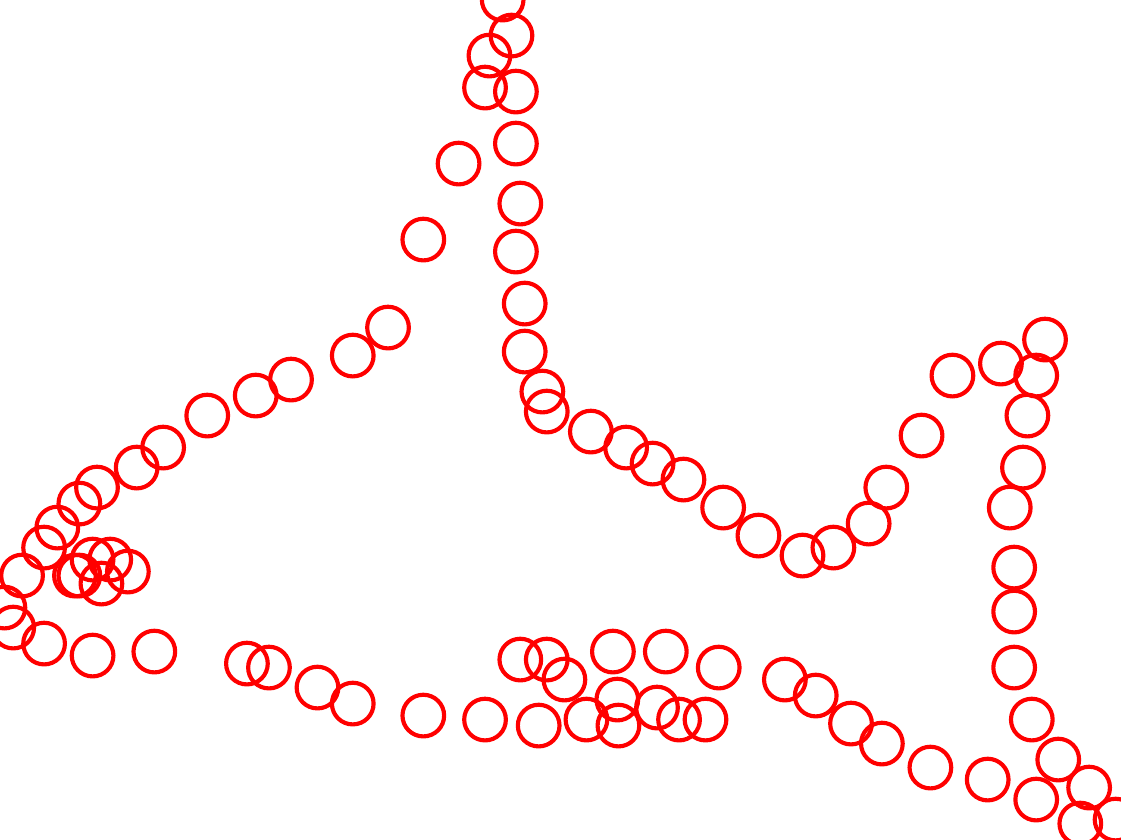}&		
		\includegraphics[width=\scale\linewidth]{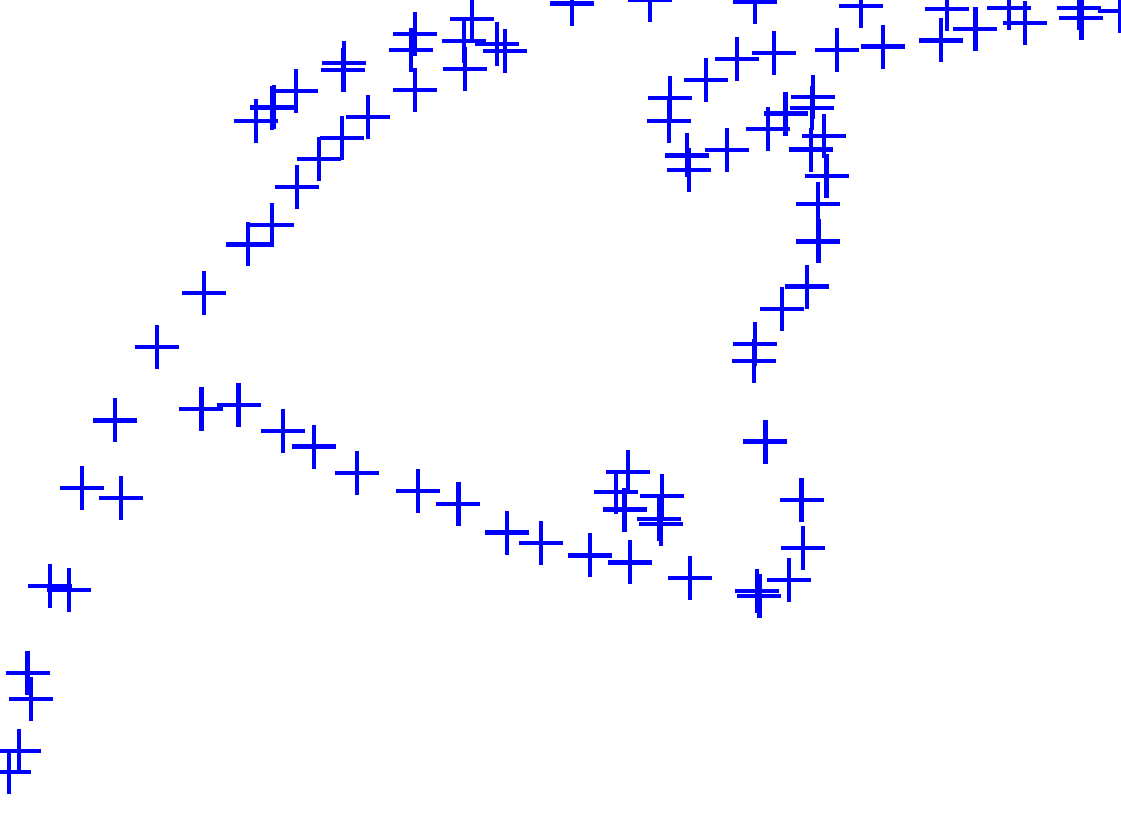}&
		\includegraphics[width=\scale\linewidth]{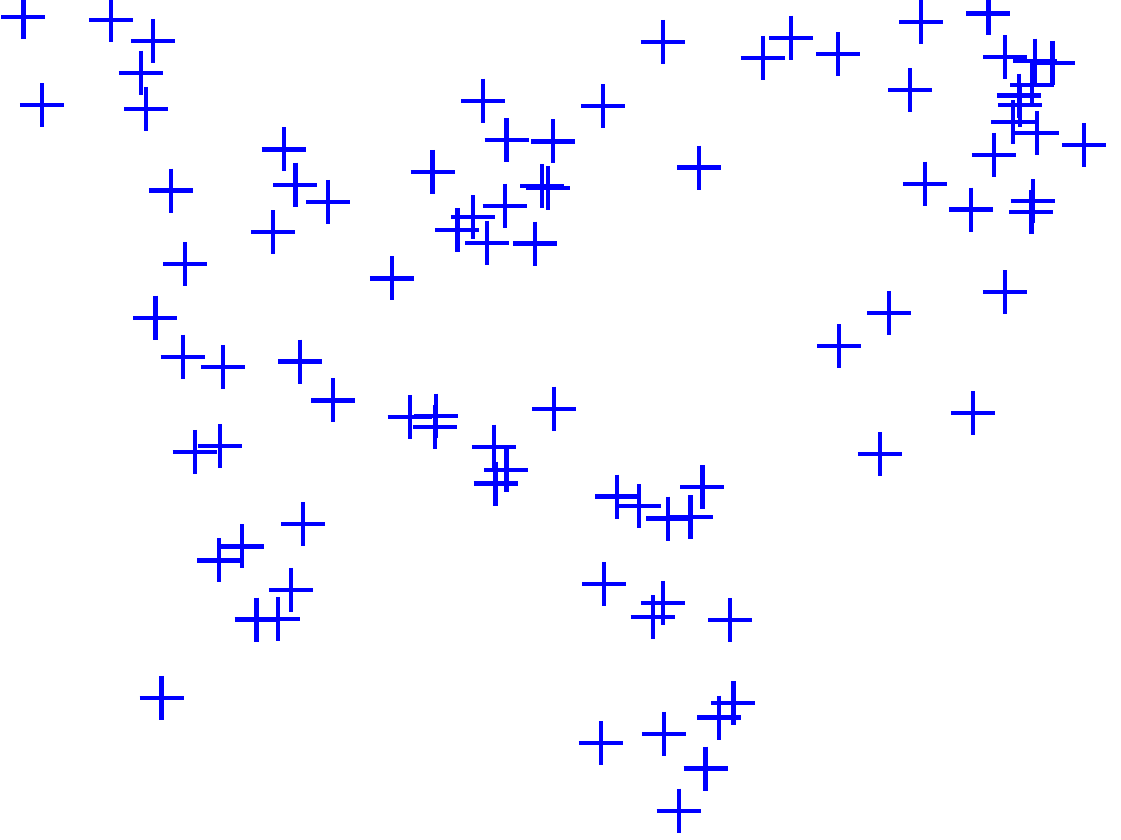}&		
		\includegraphics[width=\scale\linewidth]{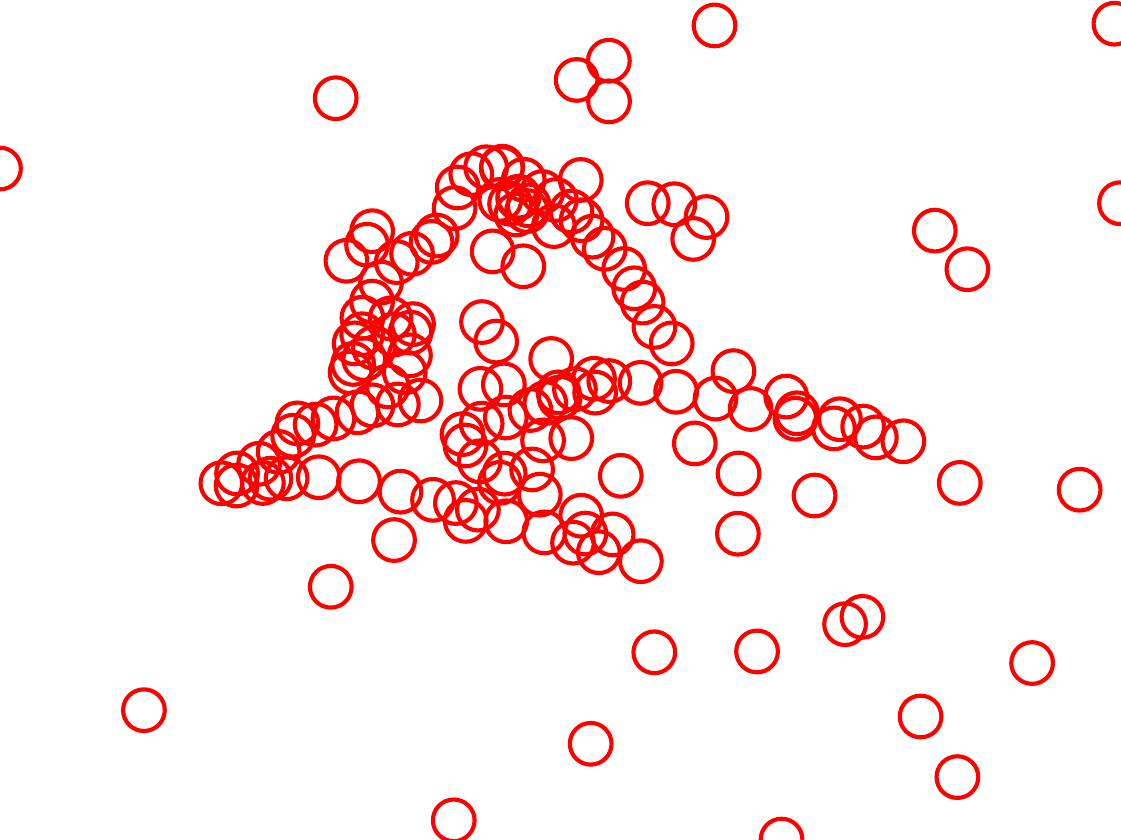}&
		\includegraphics[width=\scale\linewidth]{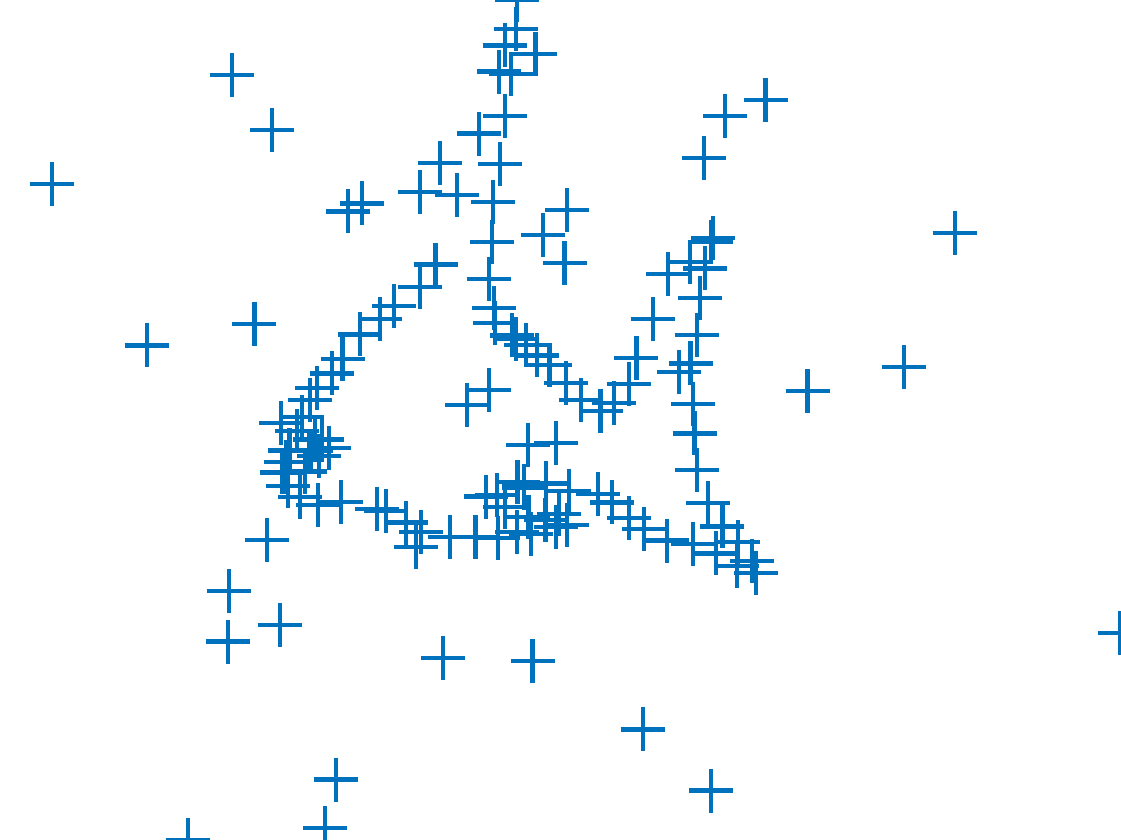}&
		\includegraphics[width=\scale\linewidth]{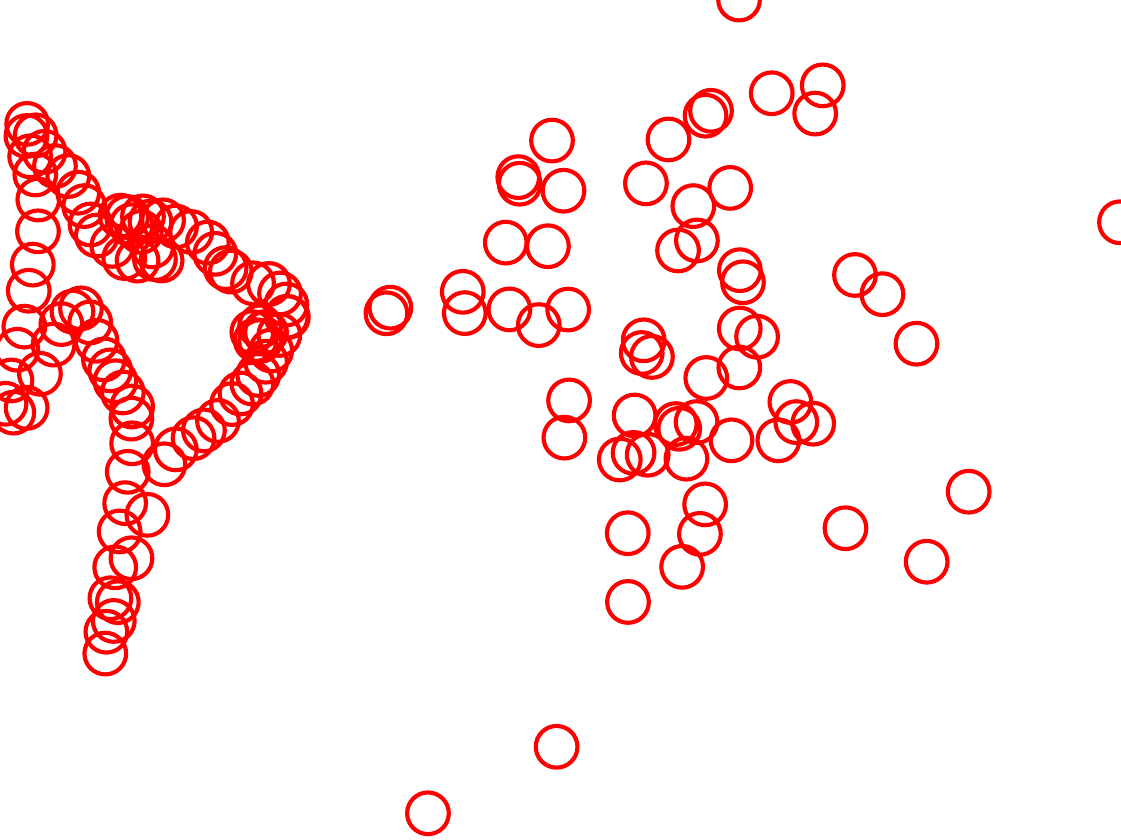}&
		\includegraphics[width=\scale\linewidth]{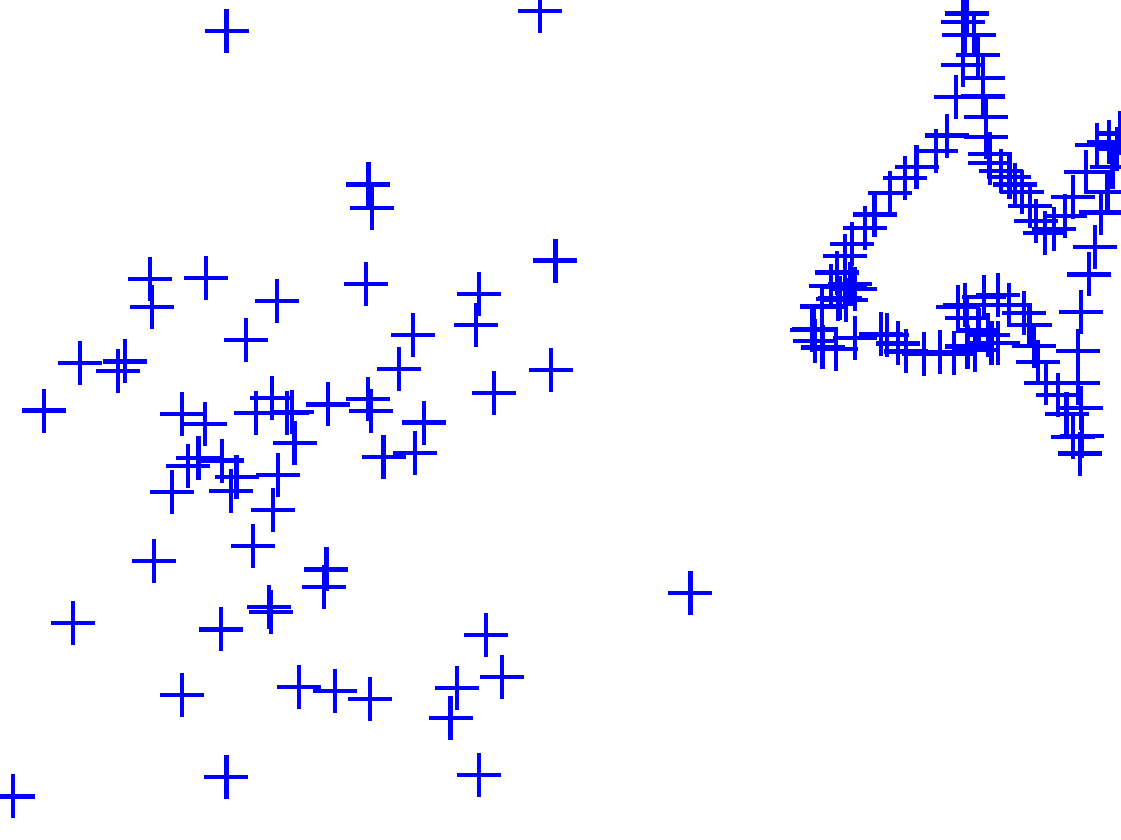}&
		\includegraphics[width=\scale\linewidth]{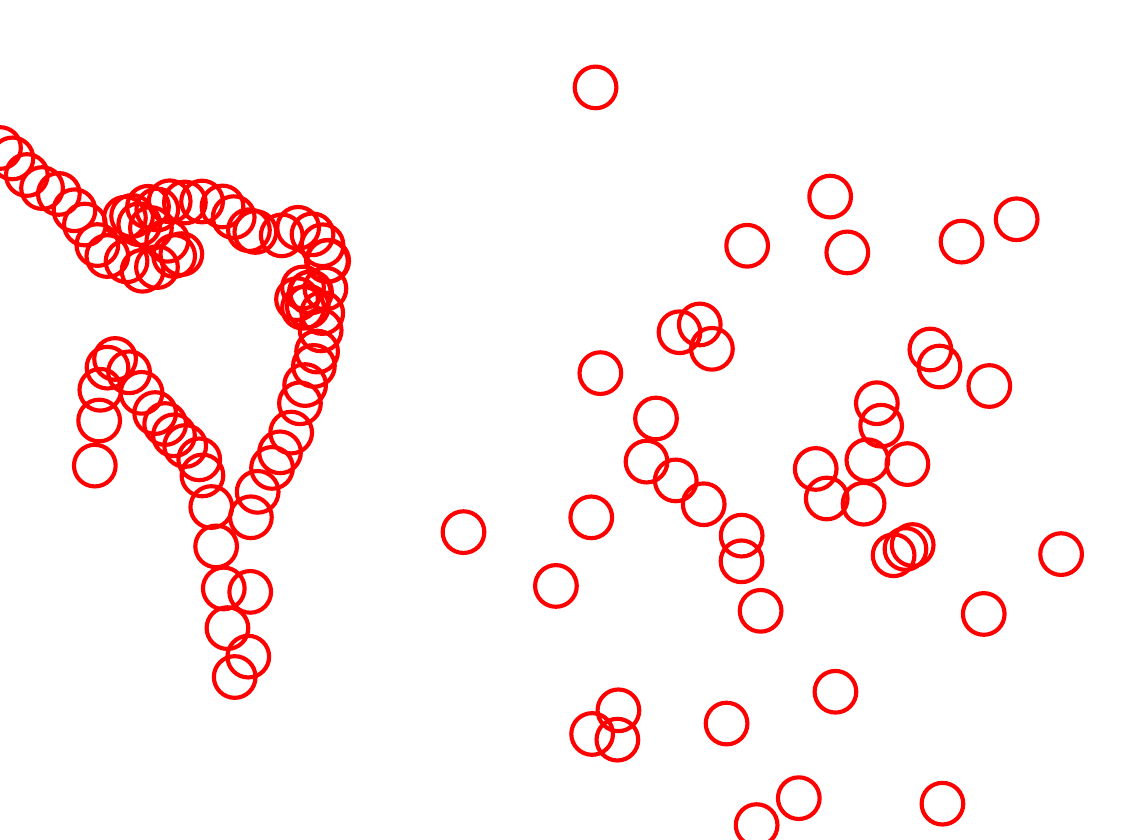}&
		\includegraphics[width=\scale\linewidth]{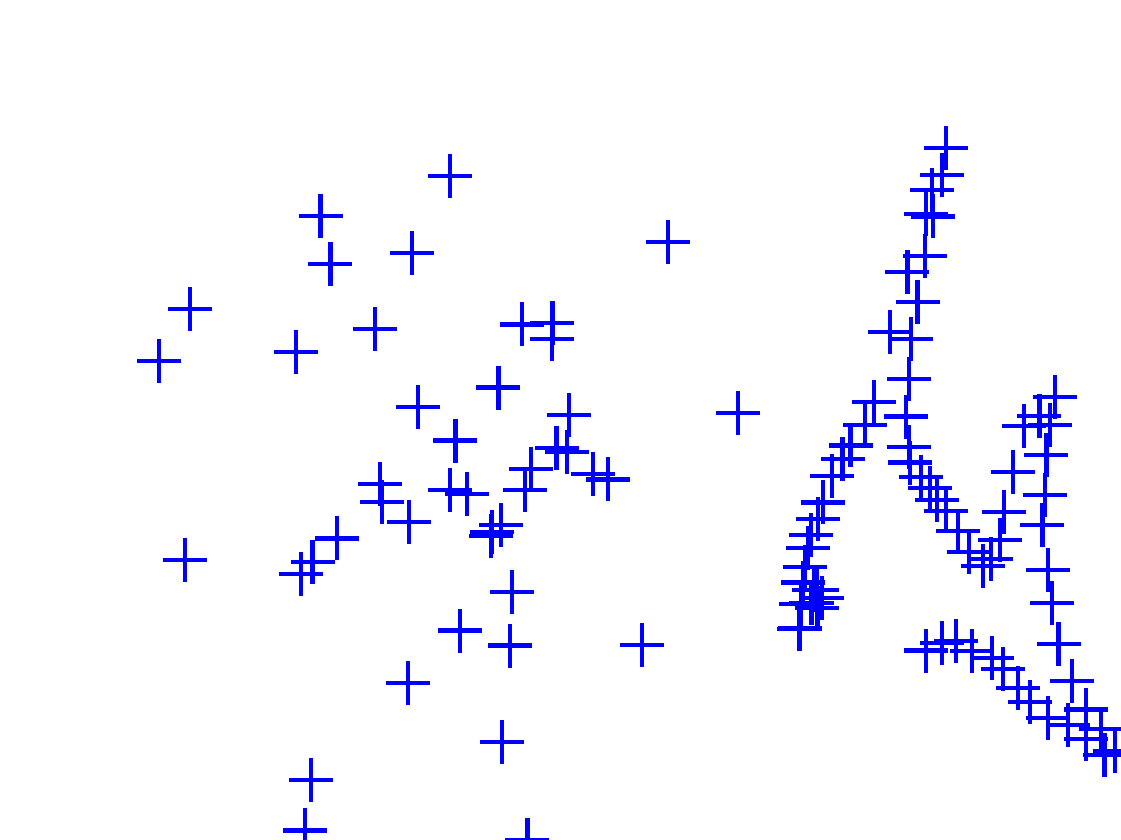} 
		\\	\hline
		{		\includegraphics[width=\scale\linewidth]{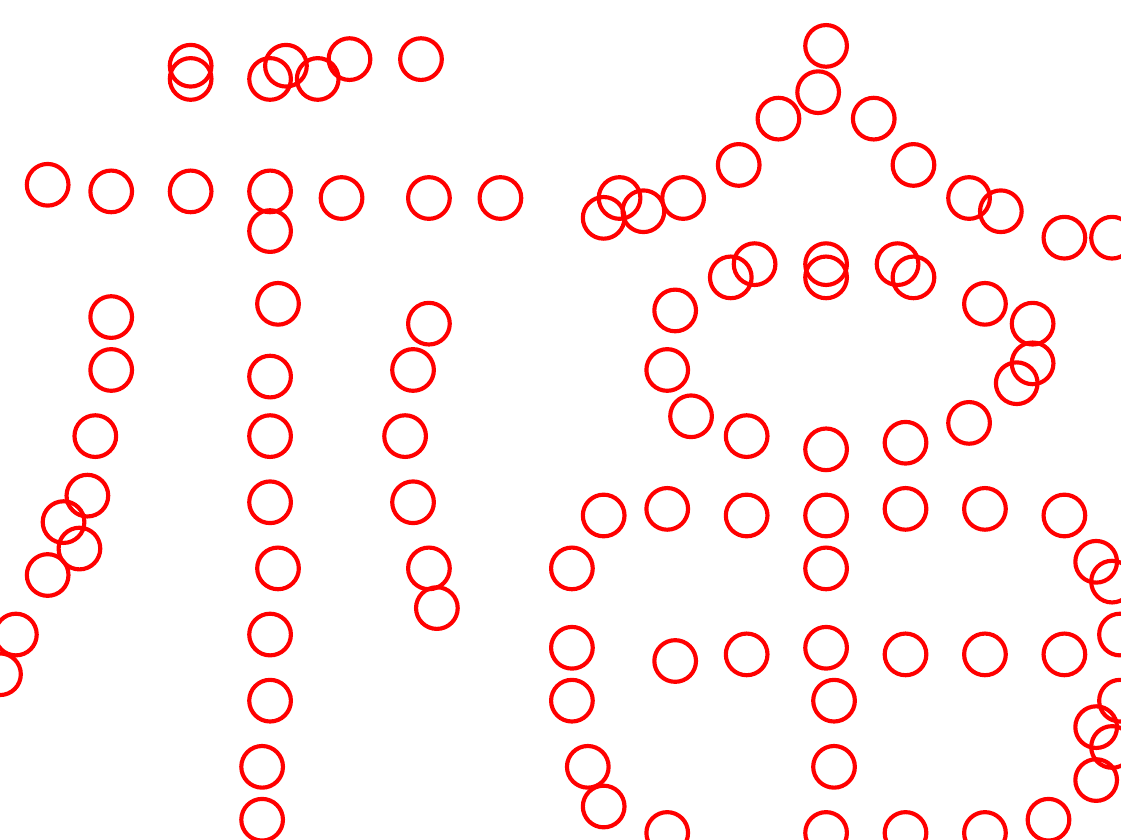}} &
		{		
			\includegraphics[width=\scale\linewidth]{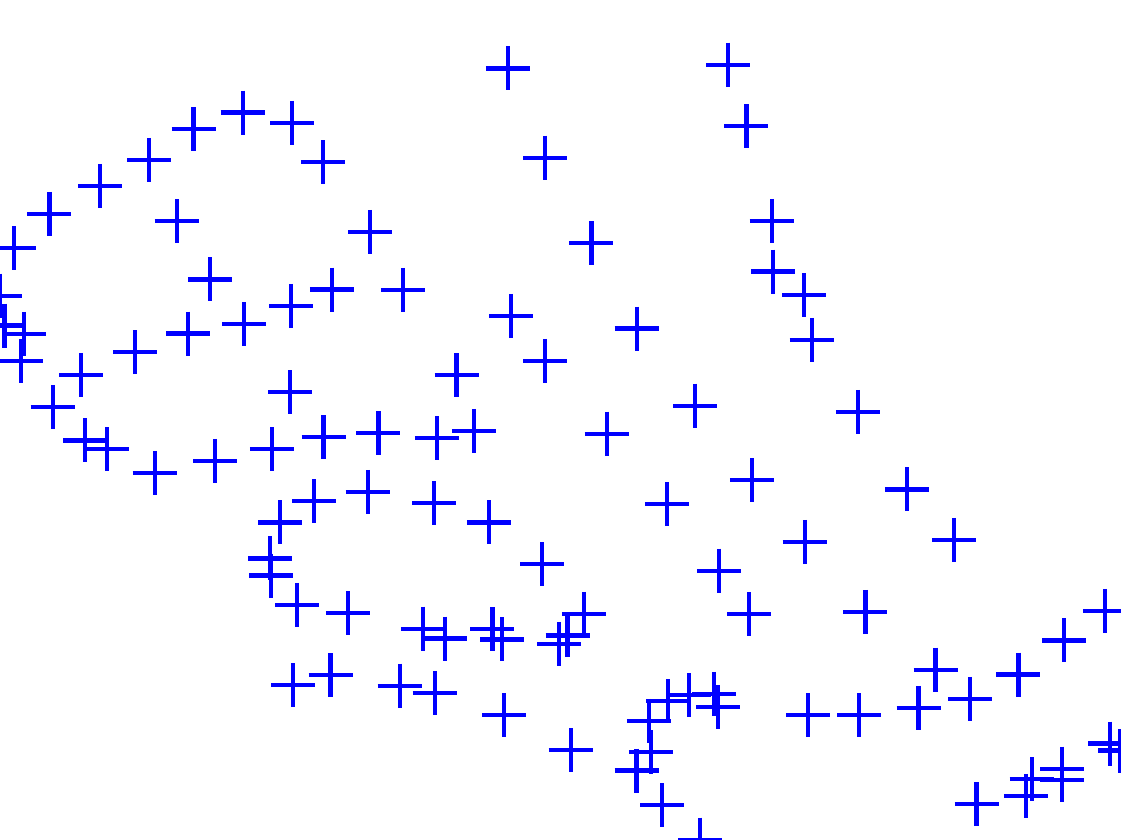}}&
		{
			\includegraphics[width=\scale\linewidth]{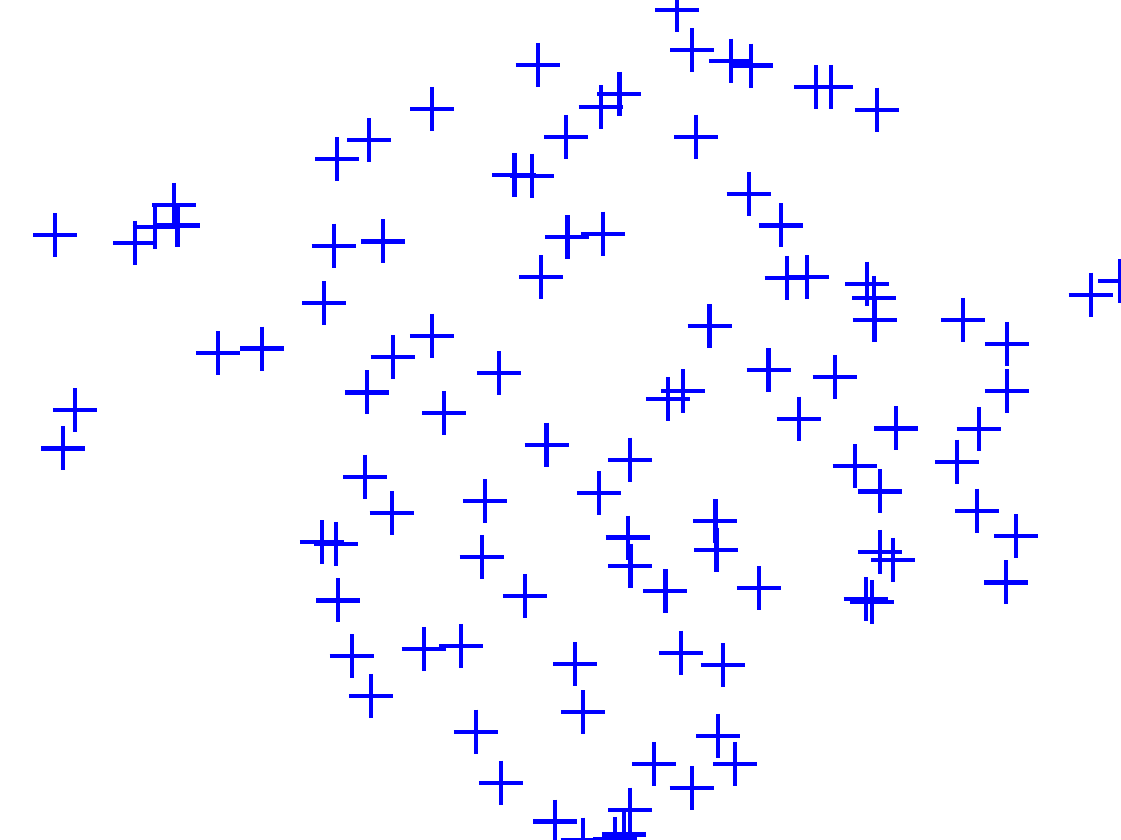}}&	{		
			\includegraphics[width=\scale\linewidth]{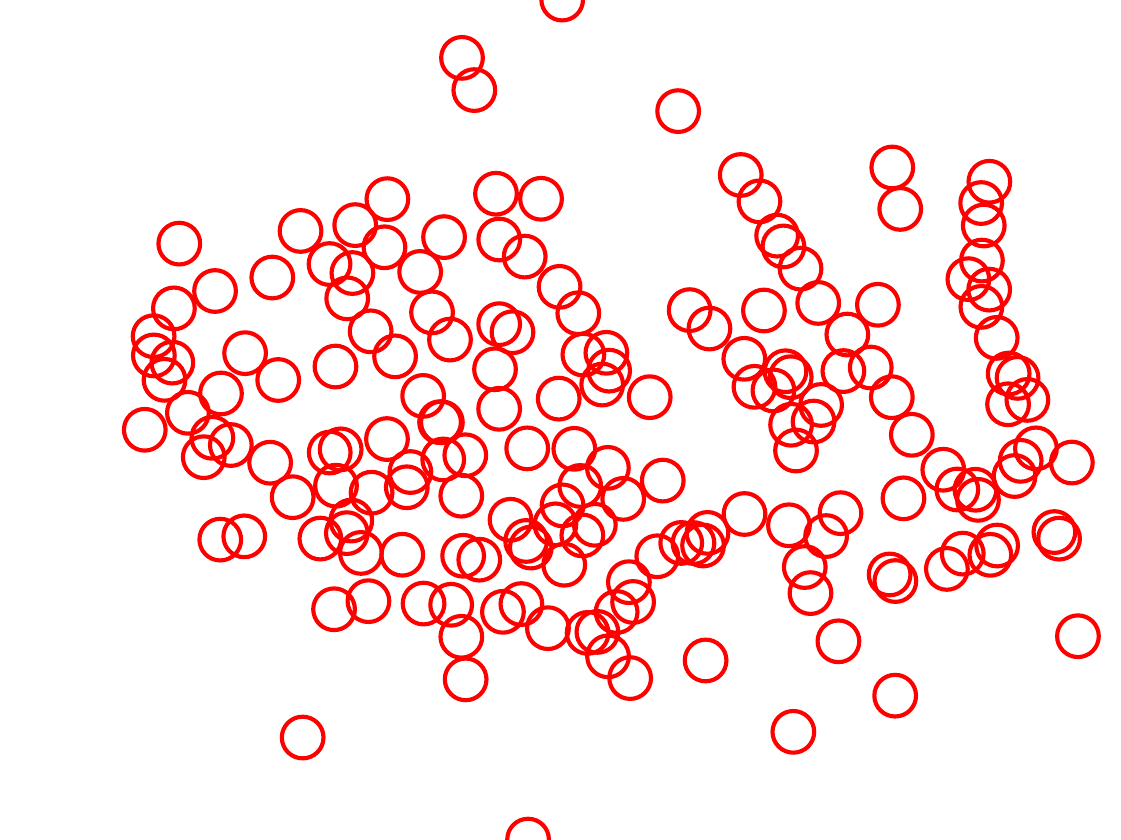}} &
		{	\includegraphics[width=\scale\linewidth]{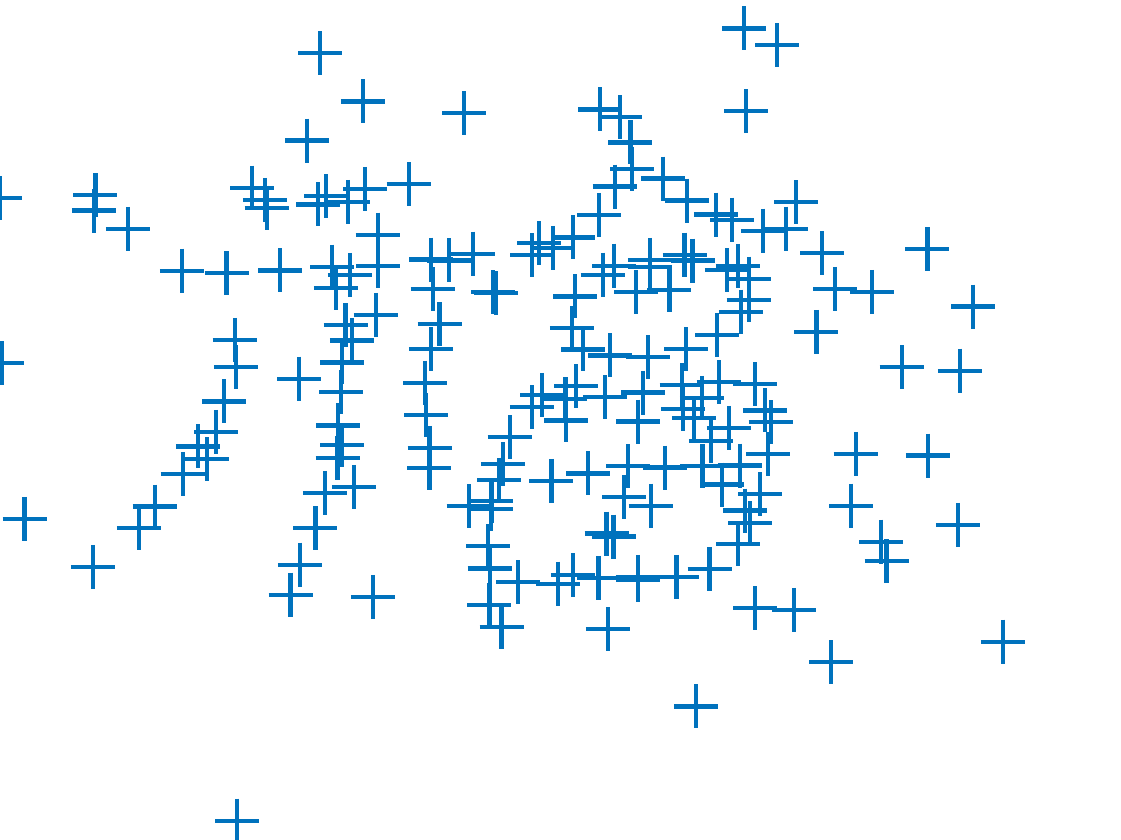}}&	  
		{ \includegraphics[width=\scale\linewidth]{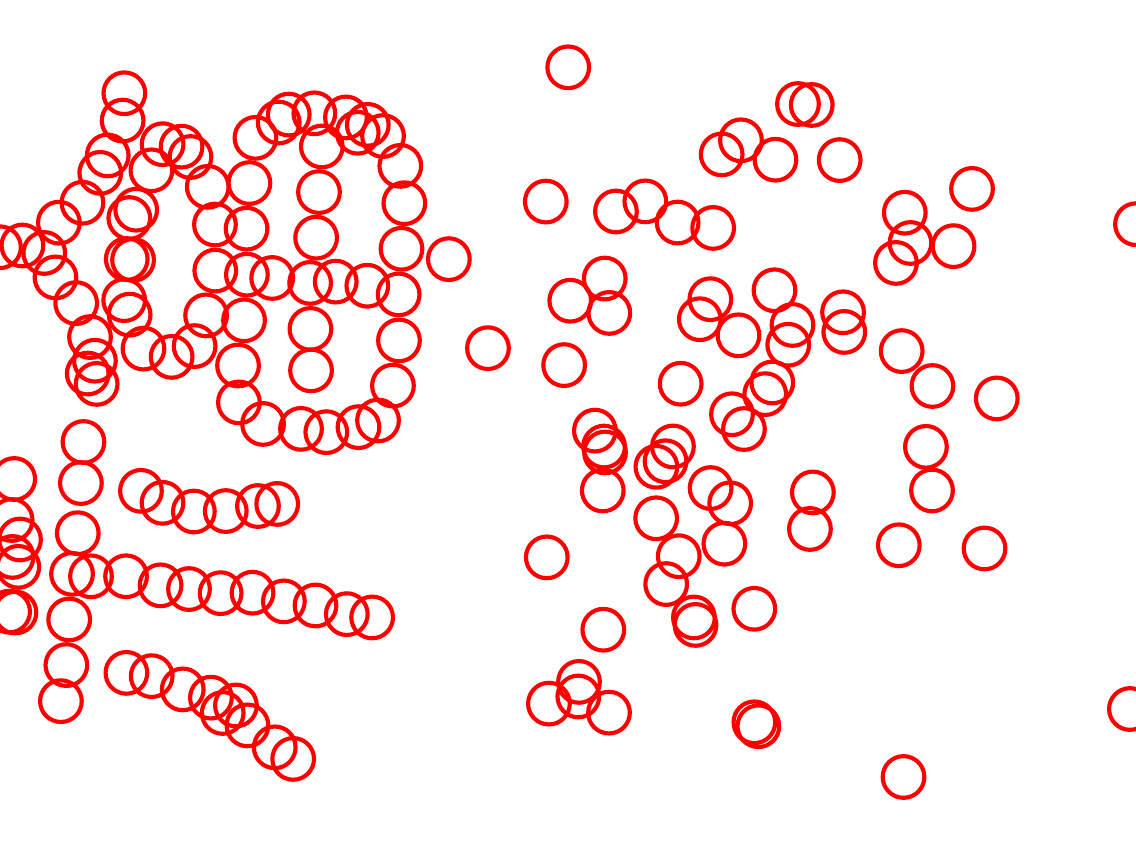}} &
		{ \includegraphics[width=\scale\linewidth]{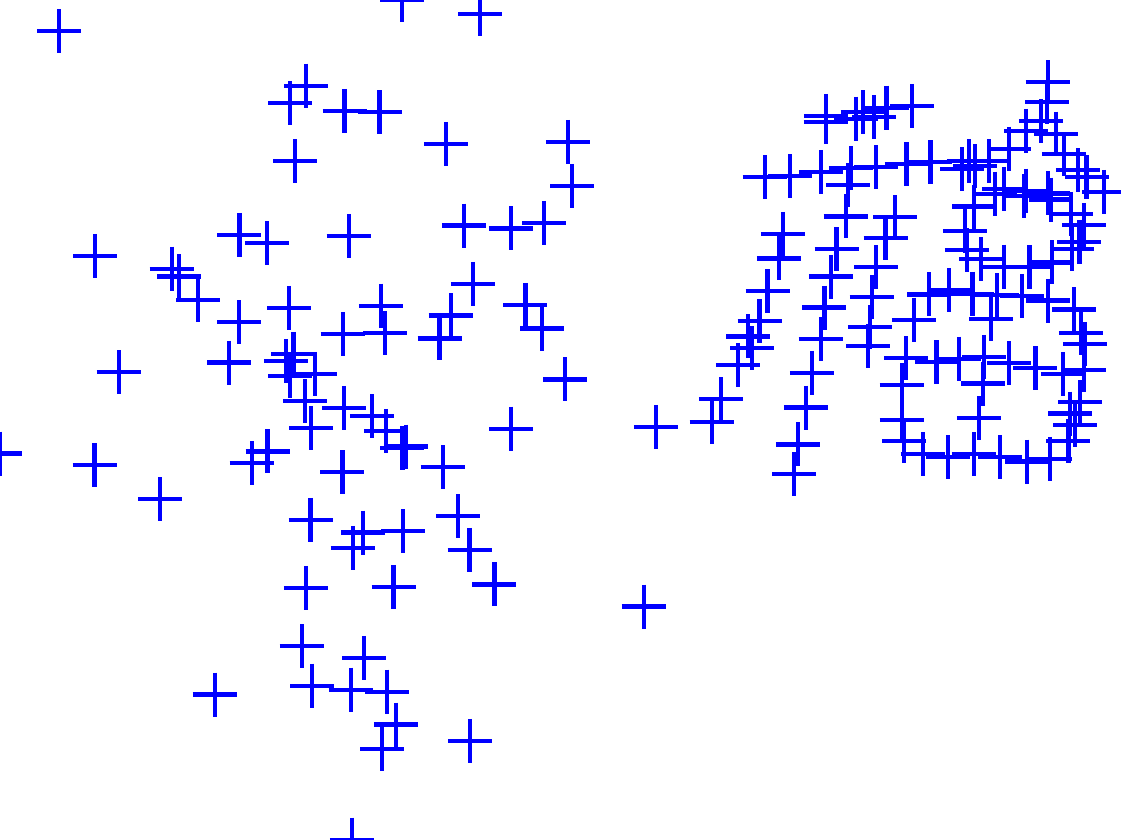}}&
		{ \includegraphics[width=\scale\linewidth]{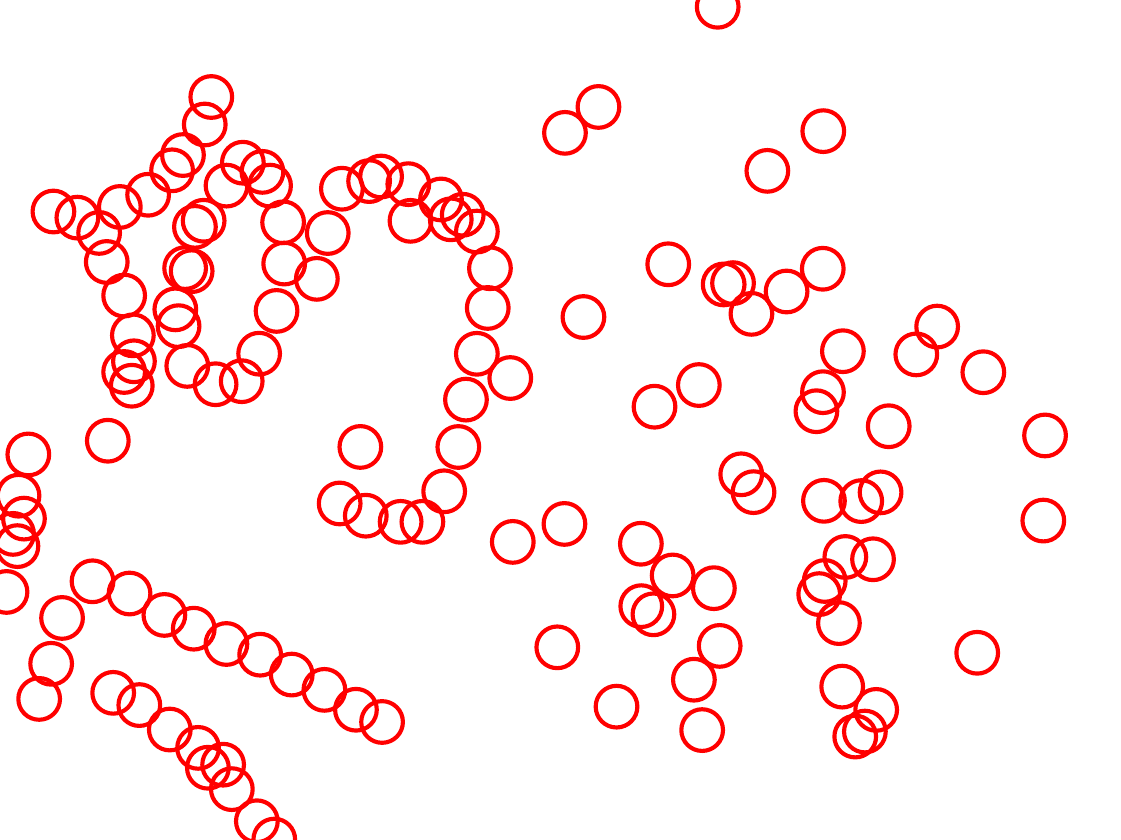}}&
		{ \includegraphics[width=\scale\linewidth]{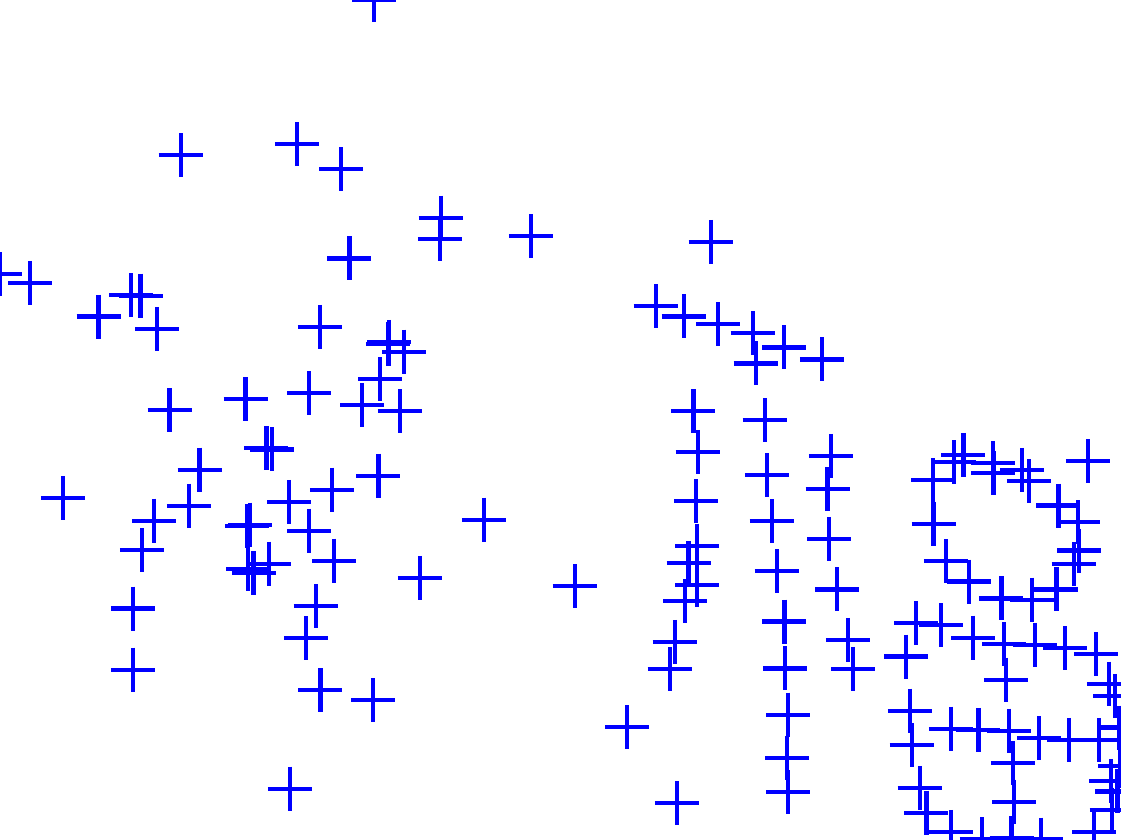}}  
	\end{tabular}
	\caption{
		a) to (c): source point sets and examples of target point sets in the deformation and noise tests, respectively.
		(d) to (i): Examples of source and target point sets in the mixed outliers and inliers test ((d), (e)), separate outliers and inliers test ((f), (g)), and occlusion+outlier test ((h), (i)), respectively.
		In all cases, source points are indicated by red circles, while scene points are represented by blue crosses.		
		\label{rot_2D_test_data_exa}}
		\centering
		\newcommand\scaleGd{10cm}
		\begin{tabular}{@{\hspace{-0mm}}c@{\hspace{-1.6mm}}c@{\hspace{-1.6mm}}c@{\hspace{-1.6mm}}c@{\hspace{-1.6mm}} c }
			\includegraphics[height=\scaleGd]{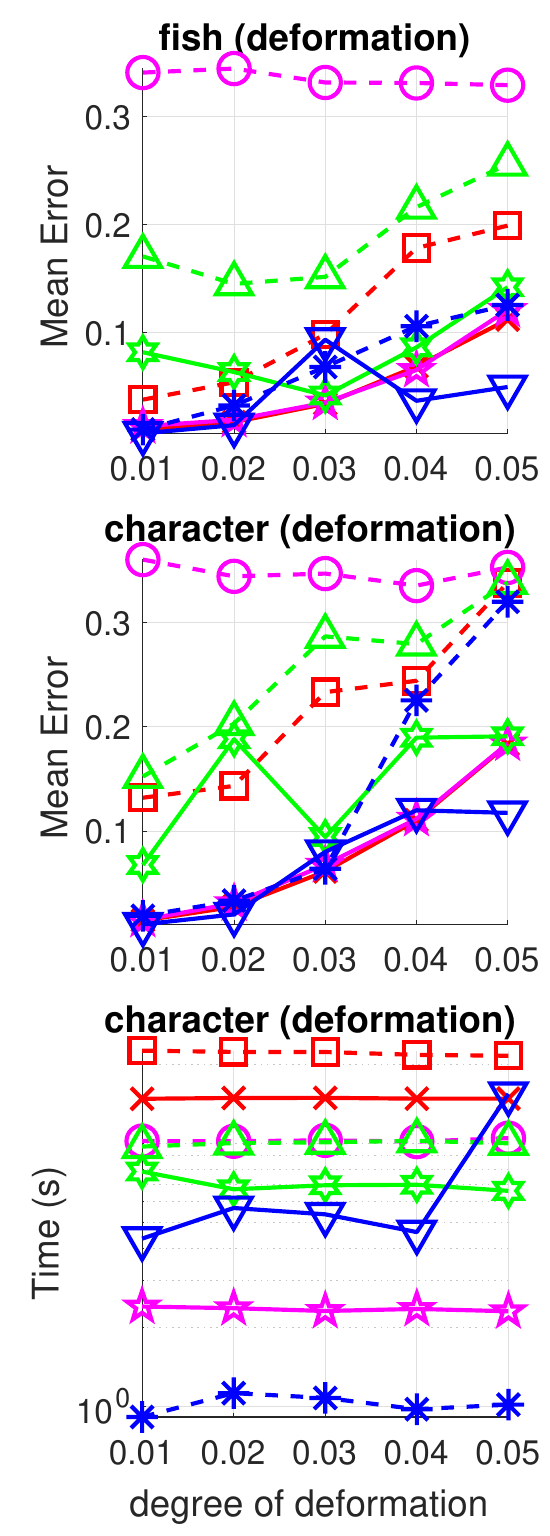}&
			\includegraphics[height=\scaleGd]{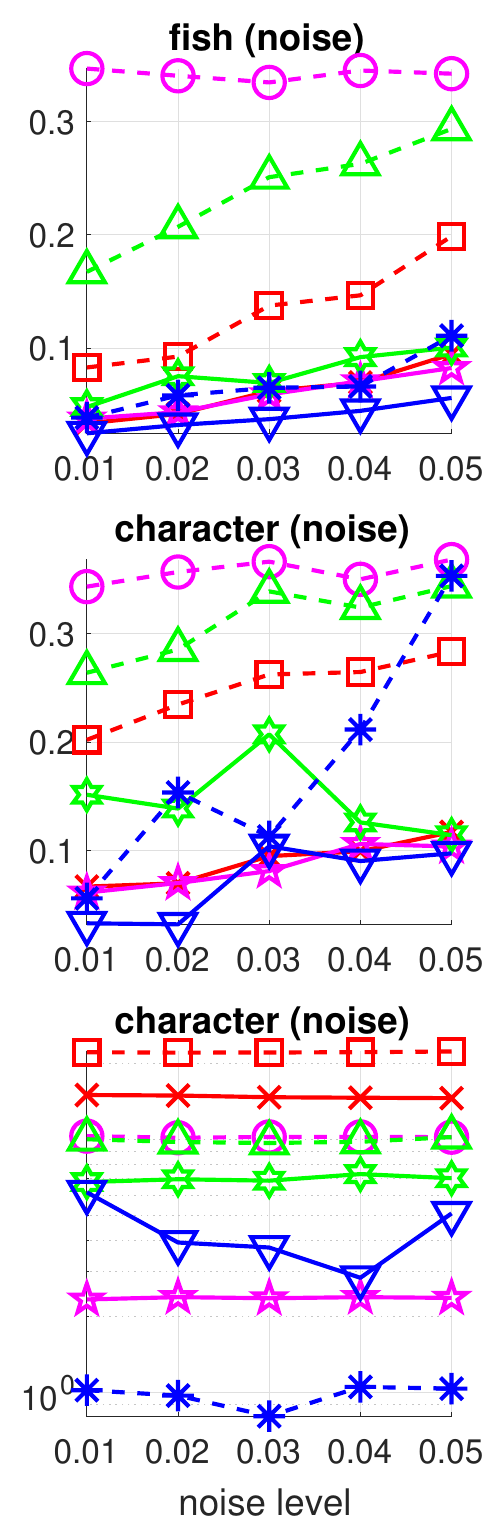}&
			\includegraphics[height=\scaleGd]{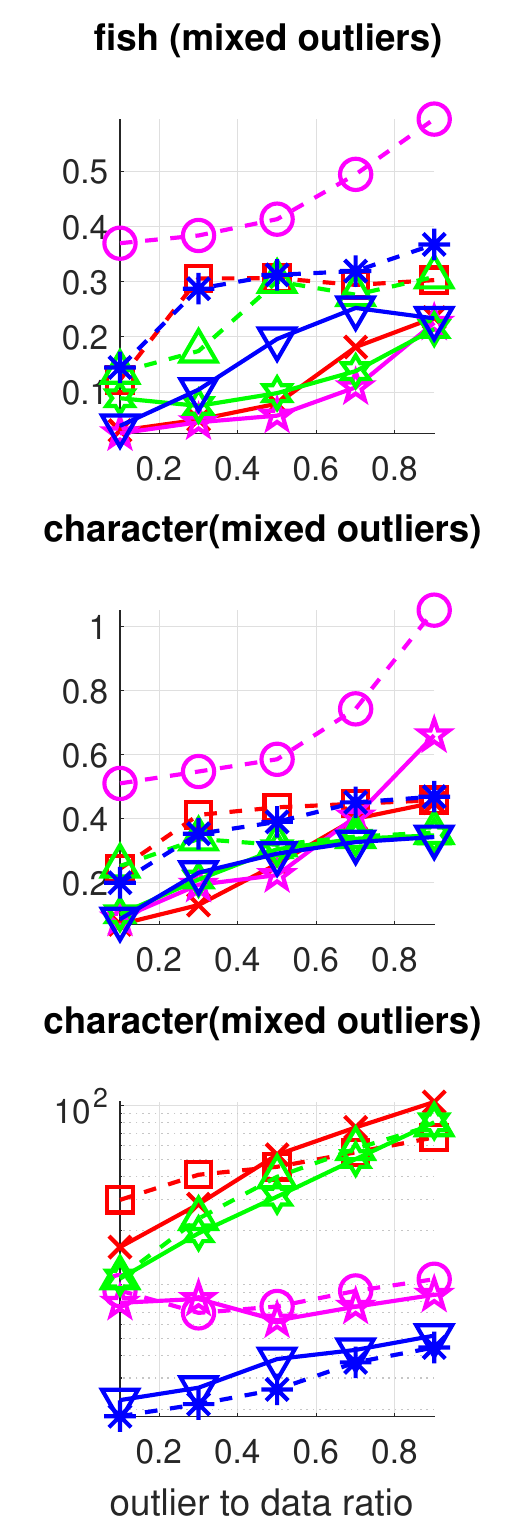}&			
			\includegraphics[height=\scaleGd]{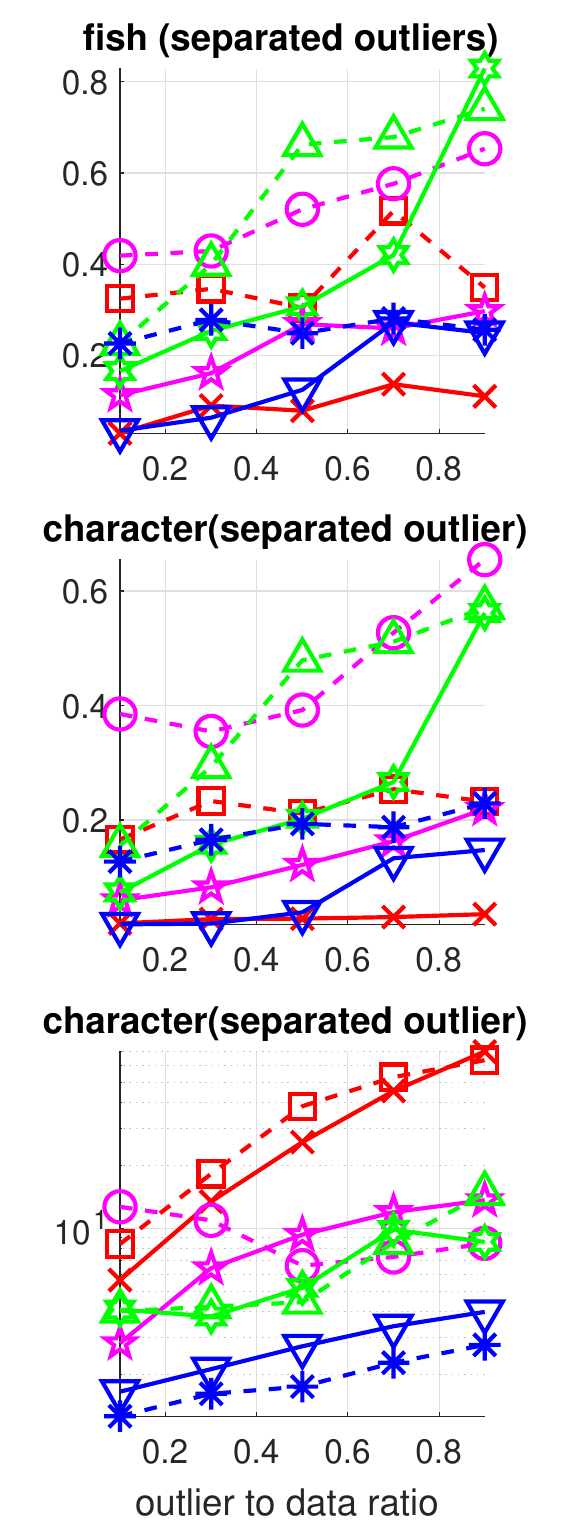}&
			\includegraphics[height=\scaleGd]{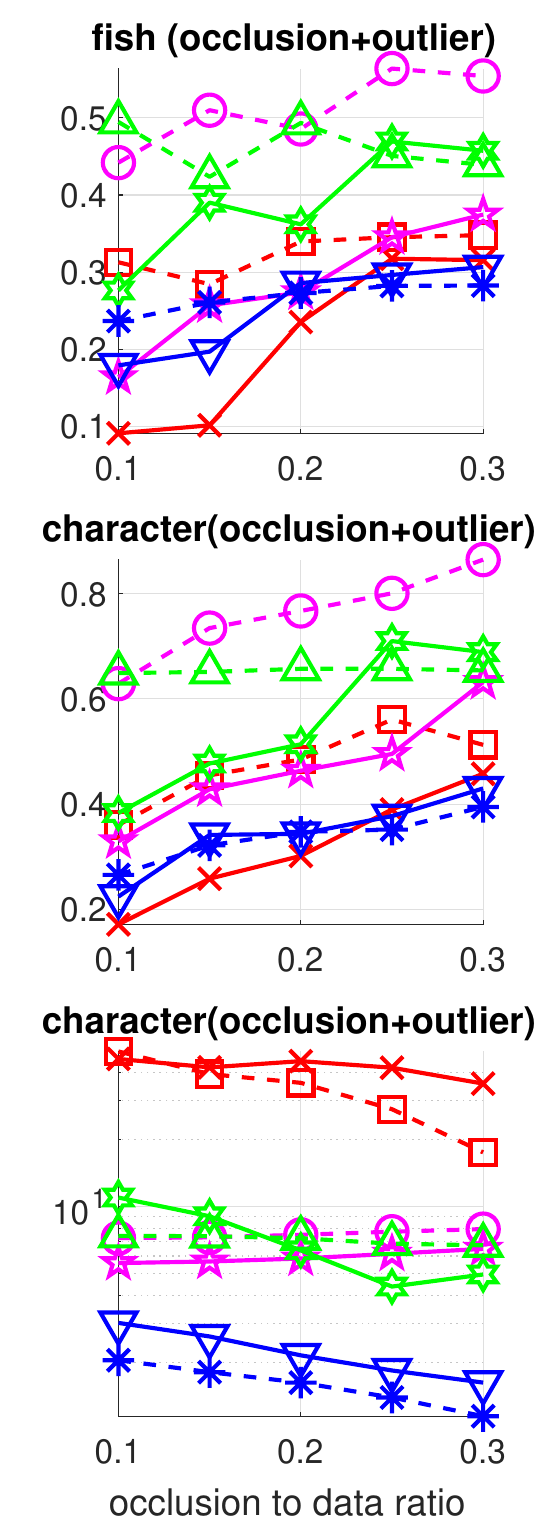}
			%
			%
		\end{tabular}
			\includegraphics[width=.7\linewidth]{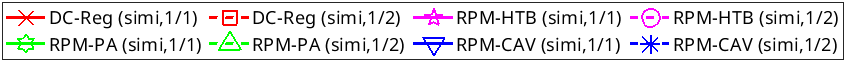}
			%
			\caption{
				Average registration errors (top 2 rows) and run times (bottom row) by 
				various methods
				under various $n_p$ values (ranging from $1/2$ to $1/1$ of the ground truth value) over 100 random trials for 2D deformation, positional noise, mixed outliers and inliers, separate outliers and inliers, and occlusion+outlier tests.
				\label{2D_simi_sta}
			}
			%
			\centering
			\includegraphics[height=5cm]{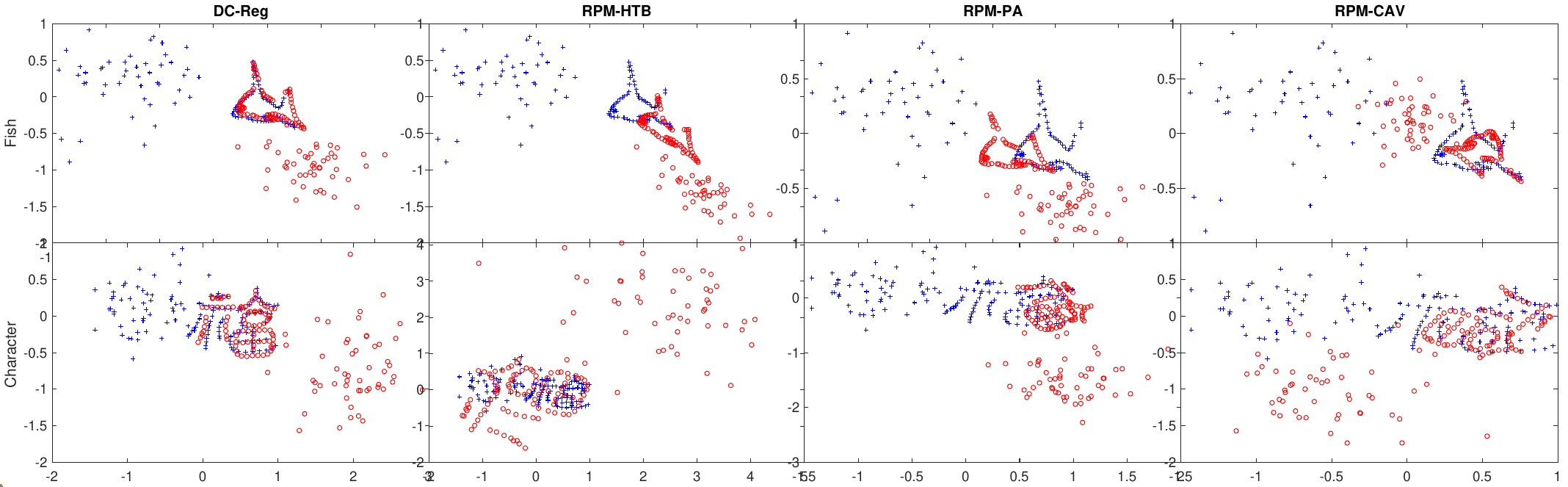}
			\caption{
				Example of registration results from different methods in the occlusion+outlier test, with $n_p$ chosen as ground truth for all methods.		
				\label{rot_2D_syn_match_exa}}	
		\end{figure*}

Using "fish" and "character" shapes, we assess the proposed algorithm's resilience to five distinct disturbances: (1) \textbf{Deformation}: Non-rigidly deforming the prototype to create the target set; (2) \textbf{Noise}: Adding positional noise to the prototype; (3) \textbf{Mixed Outliers}: Superimposing random outliers on both source and target sets; (4) \textbf{Separate Outliers}: Adding outliers to different sides of the prototype to generate distinct sets; (5) \textbf{Occlusion + Outliers}: Occluding the prototype and adding random outliers (fixed 0.5 ratio) to different sides of each set. All datasets include random rotations and scalings in the range $[0.5, 1.5]$. We evaluate two DC-Reg variants: $n_p=1/1$ (ground truth inliers) and $n_p=1/2$ (half of ground truth) to analyze parameter sensitivity.
These tests are visually illustrated in Fig. \ref{rot_2D_test_data_exa}, with alignment examples in Fig. \ref{rot_2D_syn_match_exa}. Statistical results are summarized in Fig. \ref{2D_simi_sta}.



\textbf{Robustness to Noise and Deformation.}
As shown in the top two rows of Fig. \ref{2D_simi_sta}, DC-Reg demonstrates exceptional stability in both deformation and noise tests. Notably, competing methods such as RPM-HTB and RPM-PA exhibit a sharp increase in registration error once the positional noise level exceeds 0.03. In contrast, DC-Reg maintains a low-error trajectory even under substantial deformation. Furthermore, even with an underestimated inlier count ($n_p=1/2$), our method remains highly competitive, consistently outperforming the baseline methods.

\textbf{Outlier Handling and Partial Overlap.}
In the "separated outliers" and "mixed outliers" scenarios, DC-Reg maintains near-zero error while all baseline methods scale poorly with increasing outlier ratios. This showcases the strength of our holistic bounding strategy in suppressing the influence of non-corresponding points. In the most challenging "Occlusion + Outlier" test, DC-Reg remains the most stable solver. While reducing $n_p$ to $1/2$ naturally increases the residual error, the performance still matches or exceeds the best baseline results even at high occlusion ratios ($>0.2$).

\textbf{Computational Efficiency.}
The bottom row of Fig. \ref{2D_simi_sta} illustrates the average run times. In "Deformation" and "Noise" tests, DC-Reg maintains an efficient and stable execution time (around $10^0$ to $10^1$ seconds). However, in high outlier or occlusion regimes, the run time exhibits an upward trend, reaching approximately $10^2$ seconds. This is a predictable trade-off: as the objective landscape becomes more cluttered with outliers, the Branch-and-Bound (BnB) search tree requires more iterations to prune the expanded feasible region and guarantee global optimality.

\textbf{Bounding Tightness.}
The significant performance gap between DC-Reg and RPM-HTB—which relies on loose, term-wise relaxations—validates that our holistic concave underestimator more effectively captures the coupling between transformation and correspondence variables. This theoretical advantage ensures superior global convergence, even when inlier counts are inaccurately estimated.

		\subsubsection{2D Real-World Image Registration}

\begin{figure*}[!ht]
			\centering
			\setlength\arrayrulewidth{1pt}
			
			\begin{minipage}[c]{0.1\linewidth} 
				\centering
				\includegraphics[height=8cm]{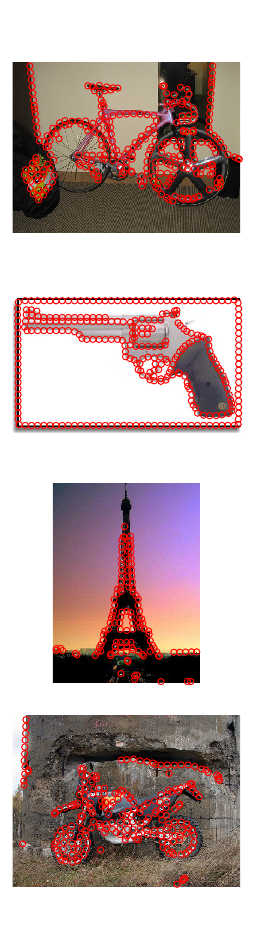}
				\vspace{2mm}\\ {\small (a)}
			\end{minipage}%
			\hspace{5mm} 
			\begin{minipage}[c]{0.69\linewidth} 
				\centering
				\begin{tabular}{c@{\hspace{0mm}}c} 
\includegraphics[height=10cm]{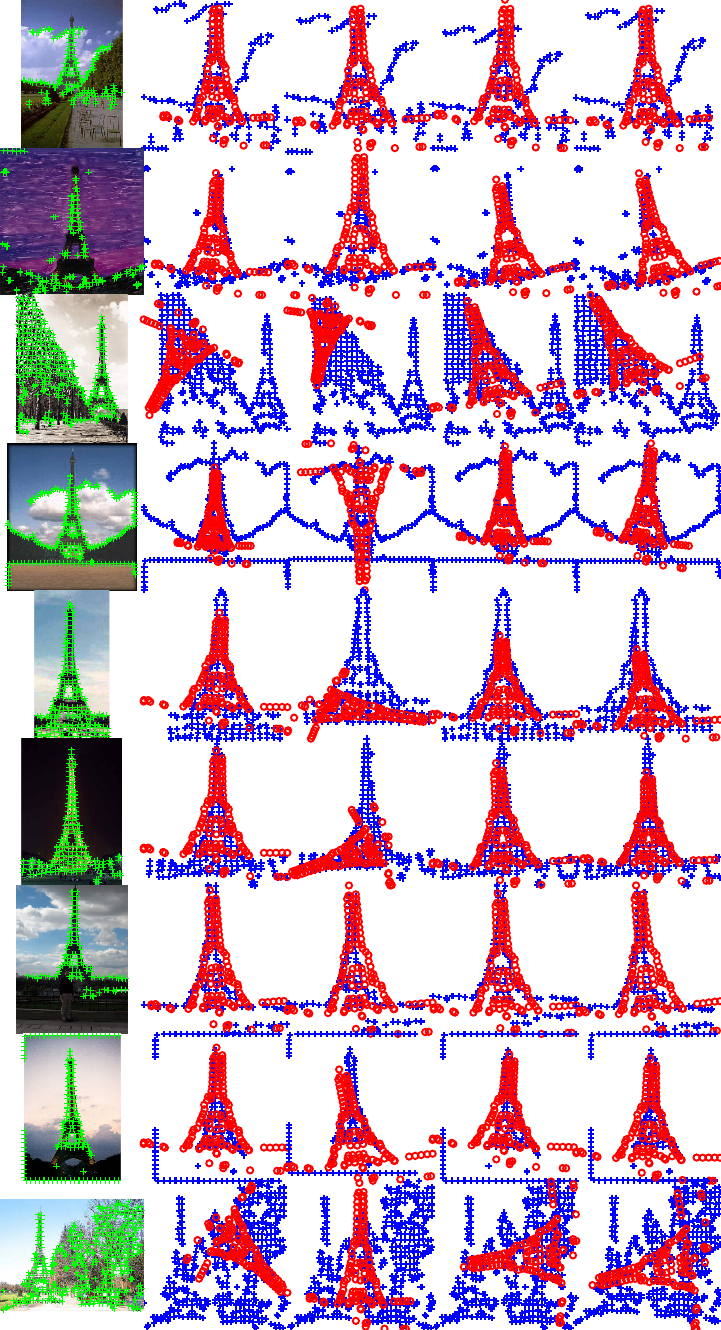} &											
\includegraphics[height=10cm]{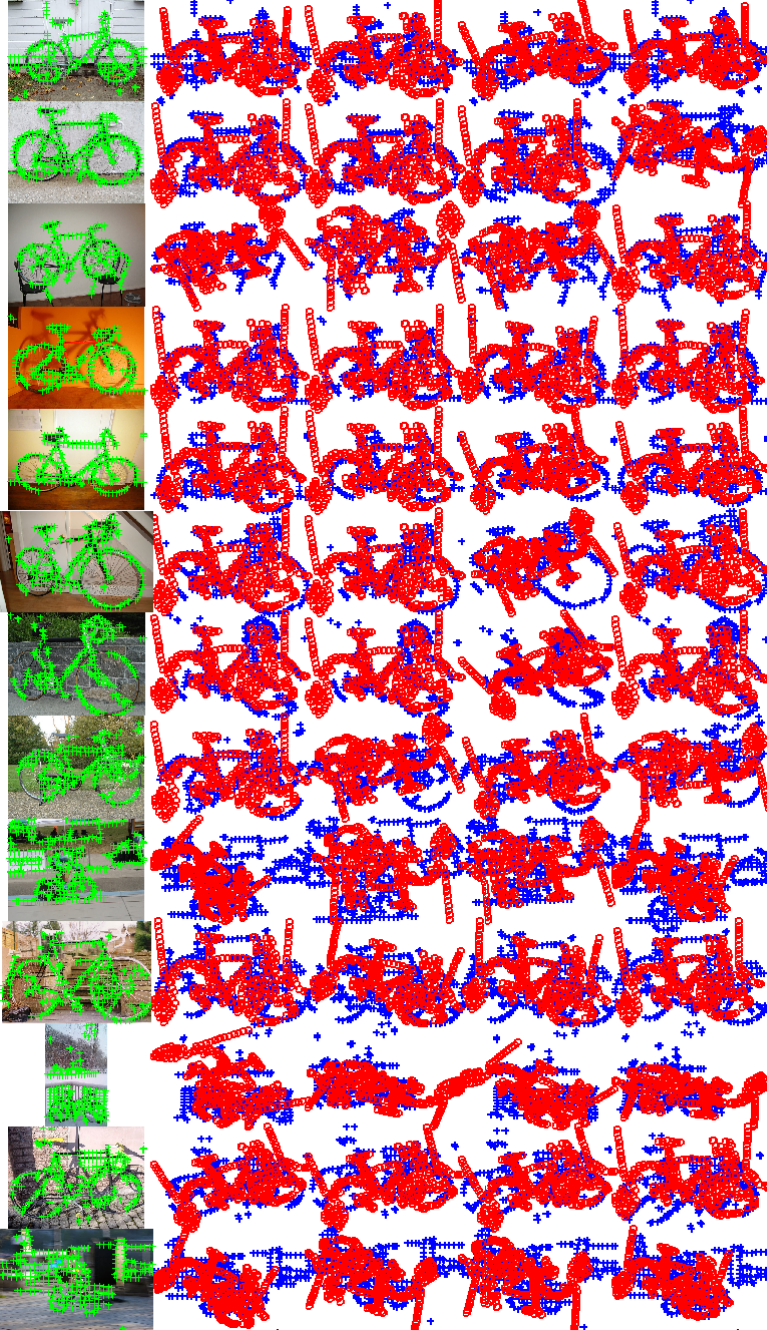} 
				\end{tabular} \\ \vspace{0mm}
				\begin{tabular}{c@{\hspace{0mm}}c}							
\includegraphics[height=8cm]{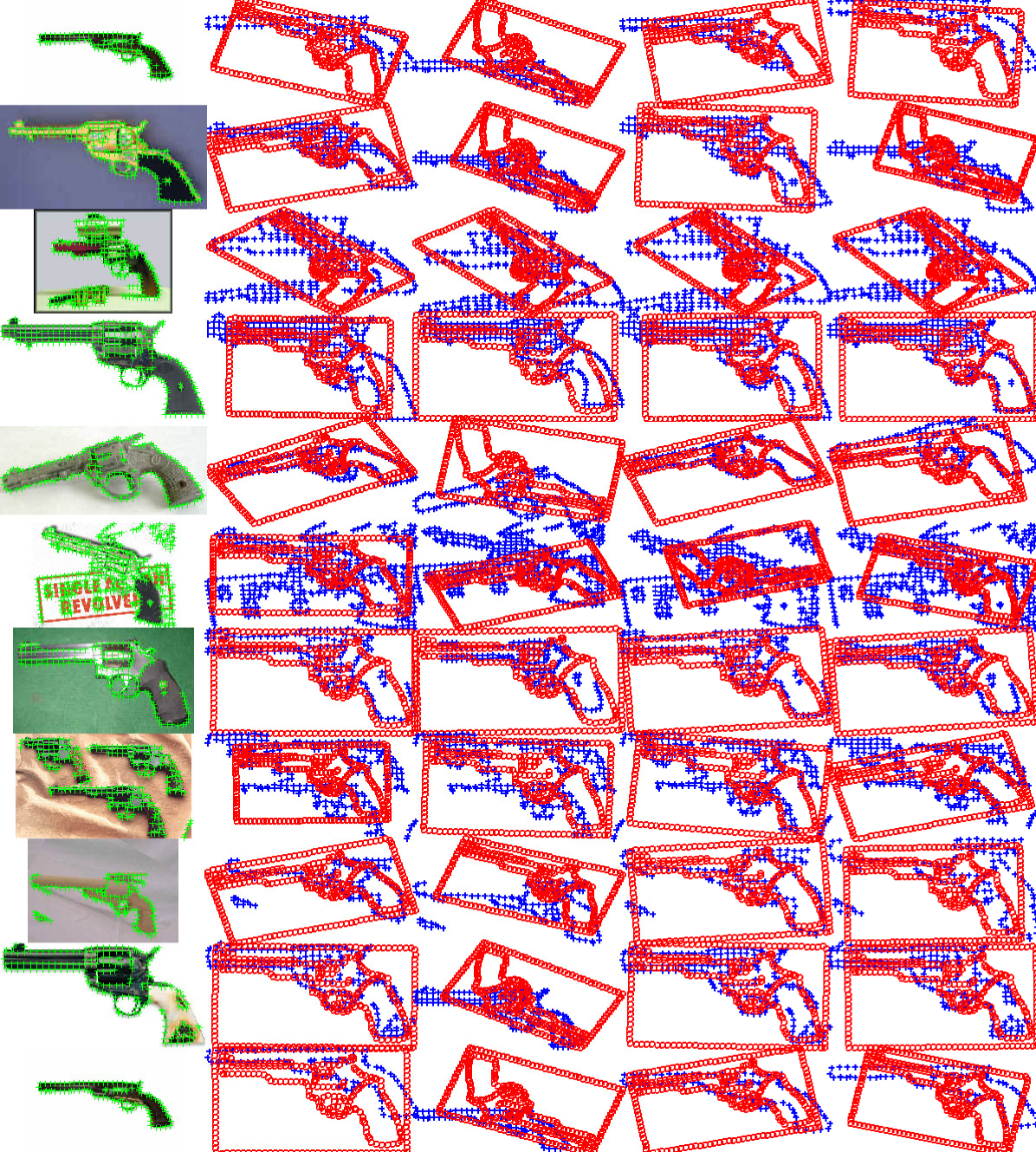} 
\includegraphics[height=8cm]{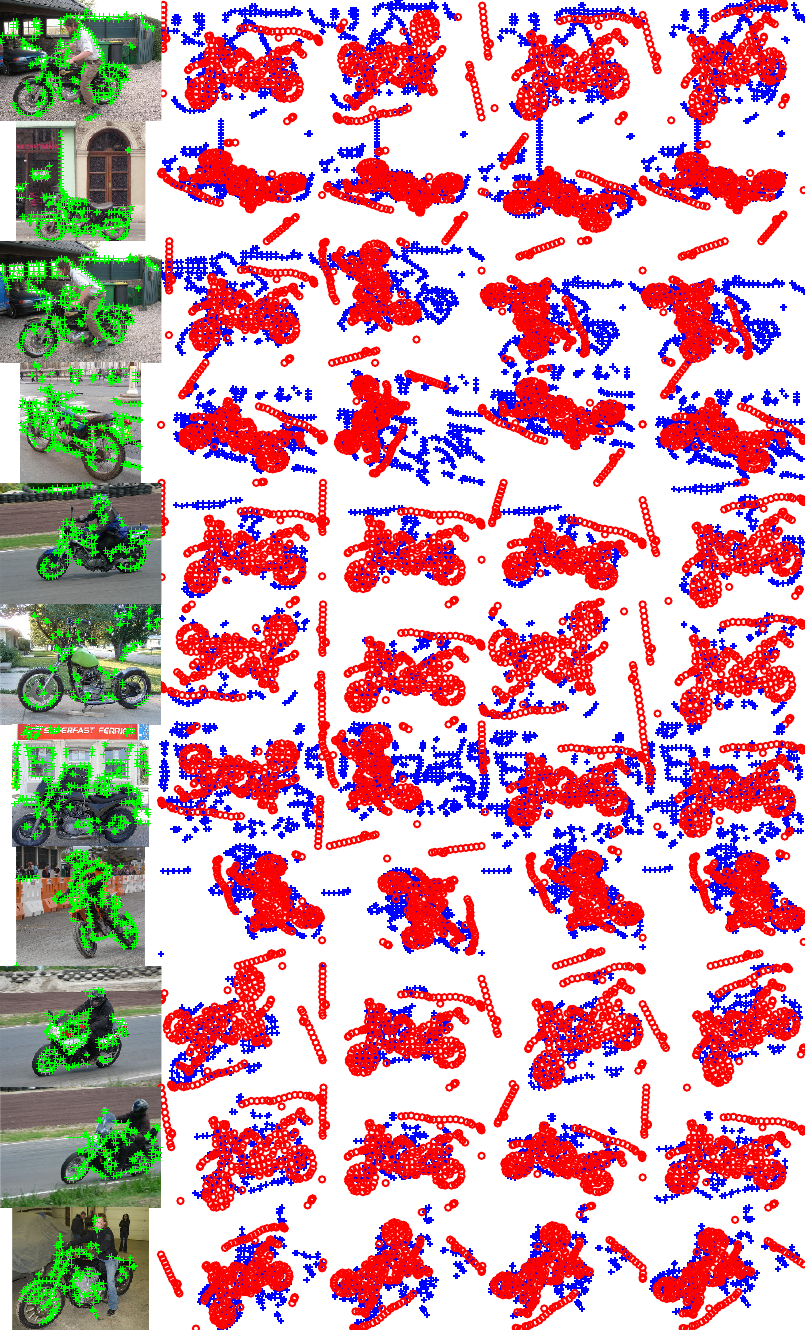} 	
				\end{tabular}
				\vspace{2mm}\\ {\small (b)}
			\end{minipage}
			
			\caption{			
				(a) 
				Source  images with source point sets superimposed.		
				(b)		For each category: 
				target images with target point sets superimposed, registration results by DC-Reg,  RPM-HTB \cite{LIAN2023126482}, RPM-PA \cite{lian2021polyhedral} and RPM-CAV \cite{RPM_model_occlude_PR} 
				using similarity  transformation.
				The $n_p$ value for each method is chosen as $0.9$ the minimum of the cardinalities of two point sets.
				\label{rot_2D_canny}}	
		\end{figure*}

		Point sets extracted from natural images provide a realistic and demanding testbed for evaluating registration robustness, as they inherently contain noise and complex structural features. We generate point sets using the Canny edge detector on images sourced from the Caltech-256 \cite{caltech_database} and VOC2007 \cite{pascal-voc-2007} datasets. To rigorously evaluate rotational invariance, each source point set is subjected to a $180^\circ$ rotation prior to registration. The source sets are visualized in  Fig. \ref{rot_2D_canny} (a), while the target sets and corresponding alignment results are presented in  Fig. \ref{rot_2D_canny} (b).

		As illustrated in Fig. \ref{rot_2D_canny} (b), DC-Reg achieves superior performance in complex scenarios: 
		\begin{inparaenum}[\upshape i\upshape)]
			\item \textbf{Alignment Accuracy:} DC-Reg achieves near-perfect overlap between transformed source (red) and target (blue) sets. This is most pronounced in the "bicycle" tests where contours are intricate.  
			\item \textbf{Clutter and Occlusion:} In cases with heavy background clutter (e.g., motorbikes) or significant occlusion (e.g., Eiffel Tower), baselines frequently converge to incorrect local minima. DC-Reg successfully recovers the global optimum, demonstrating the effectiveness of the proposed underestimator in capturing global spatial relationships. 
		\end{inparaenum}

		\subsection{3D Registration}

		\subsubsection{Evaluation on 3D Synthetic Data \label{sec:3D_synth_test}}
\begin{figure*} [!ht]
			\setlength{\abovecaptionskip}{-1pt plus 0pt minus 2pt} 
			\centering
			\newcommand\scale{0.104}		
			\begin{tabular}{@{\hspace{-3mm}}c@{}|@{}c@{}|@{}c@{}|@{}c@{}|@{}c@{}|@{} c@{}|@{} c@{} |c@{}|c }						
				\includegraphics[width=\scale\linewidth]{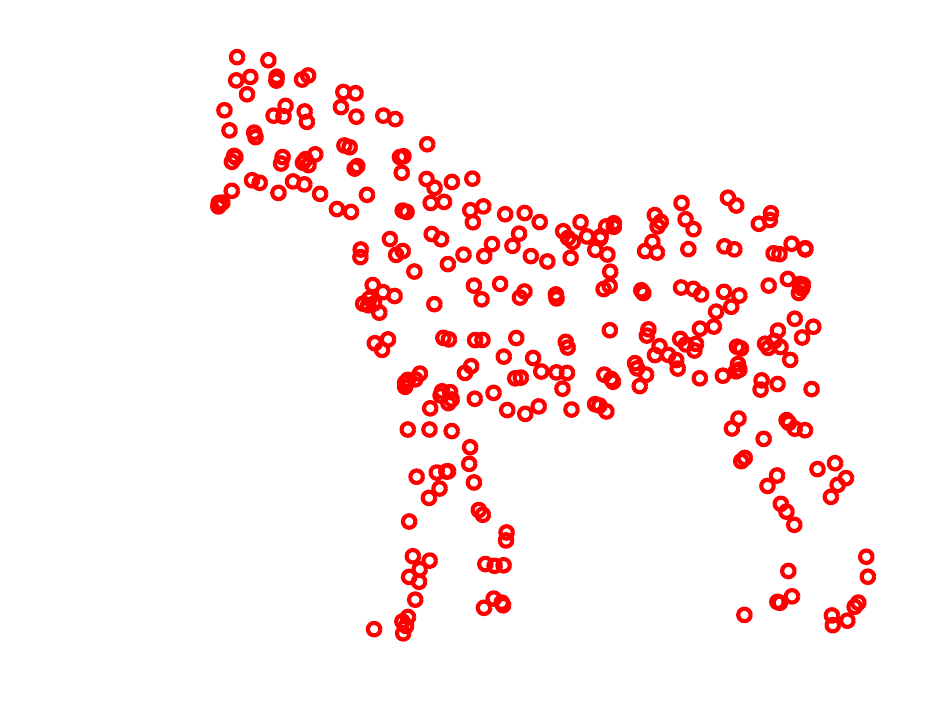}&
				\includegraphics[width=\scale\linewidth]{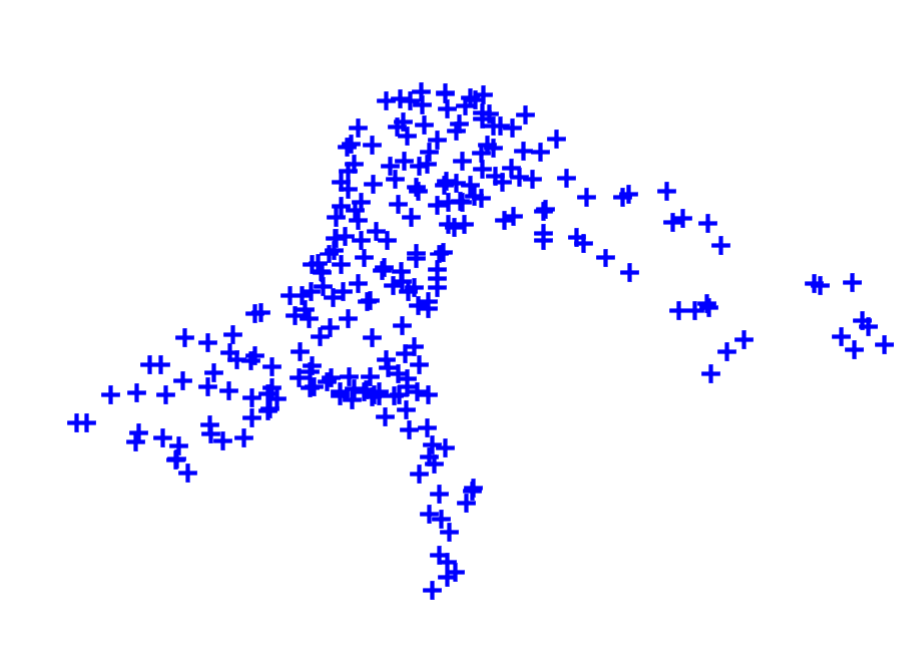}&
				\includegraphics[width=\scale\linewidth]{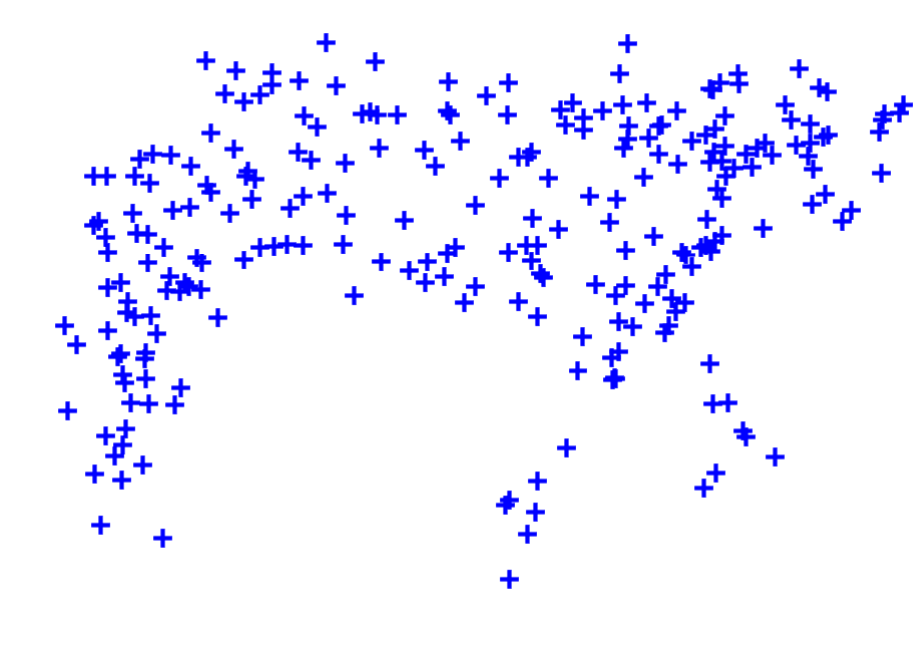}&
				\includegraphics[width=\scale\linewidth]{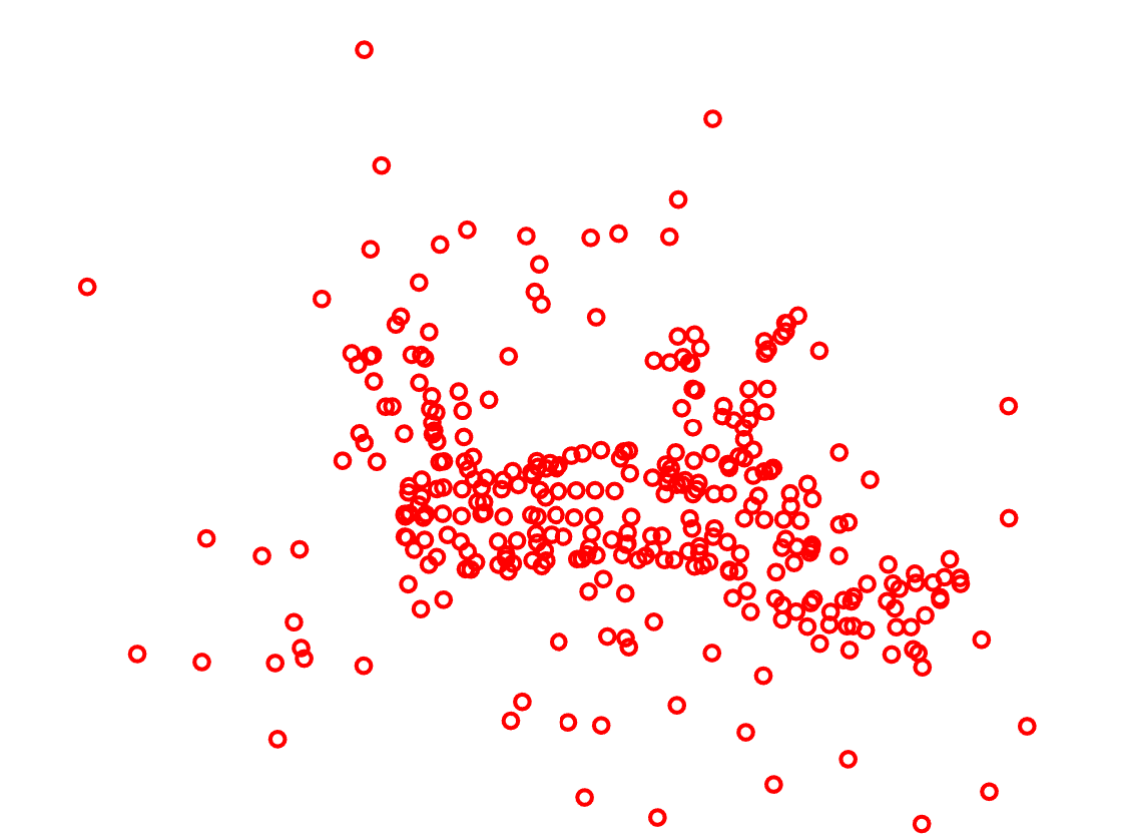}&	
				\includegraphics[width=\scale\linewidth]{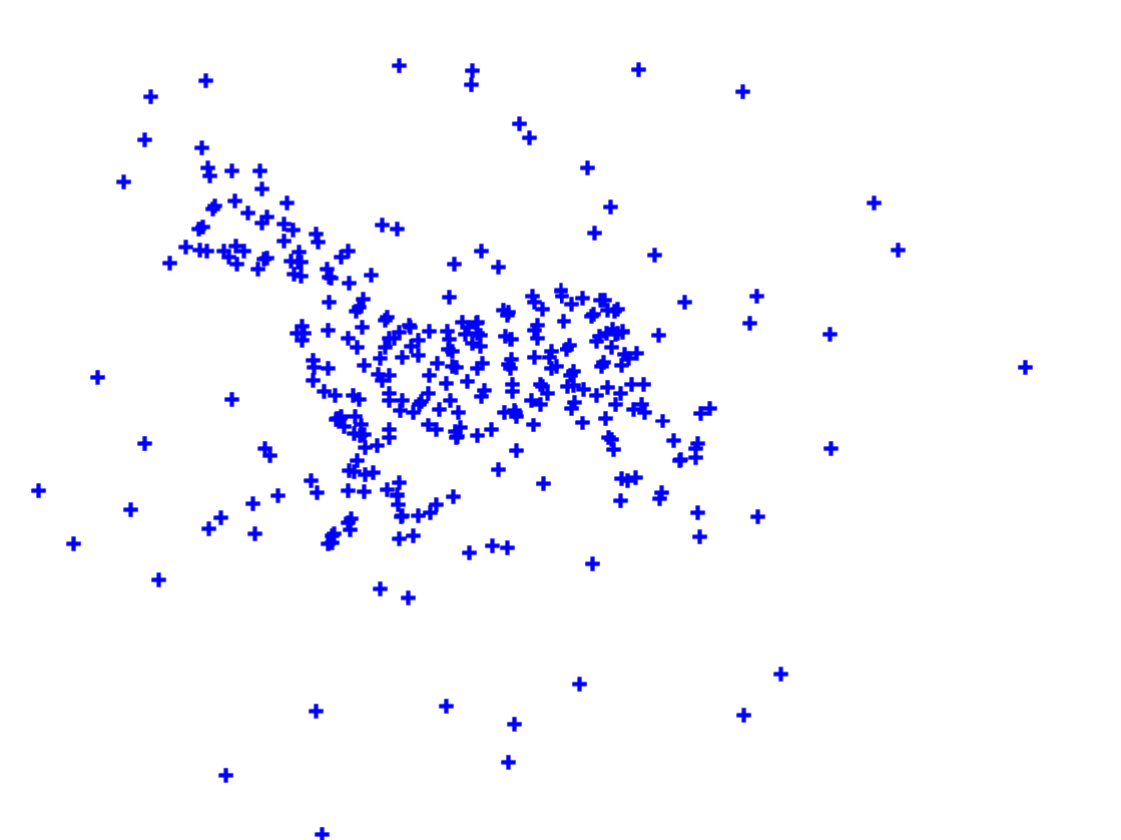}&					
				\includegraphics[width=\scale\linewidth]{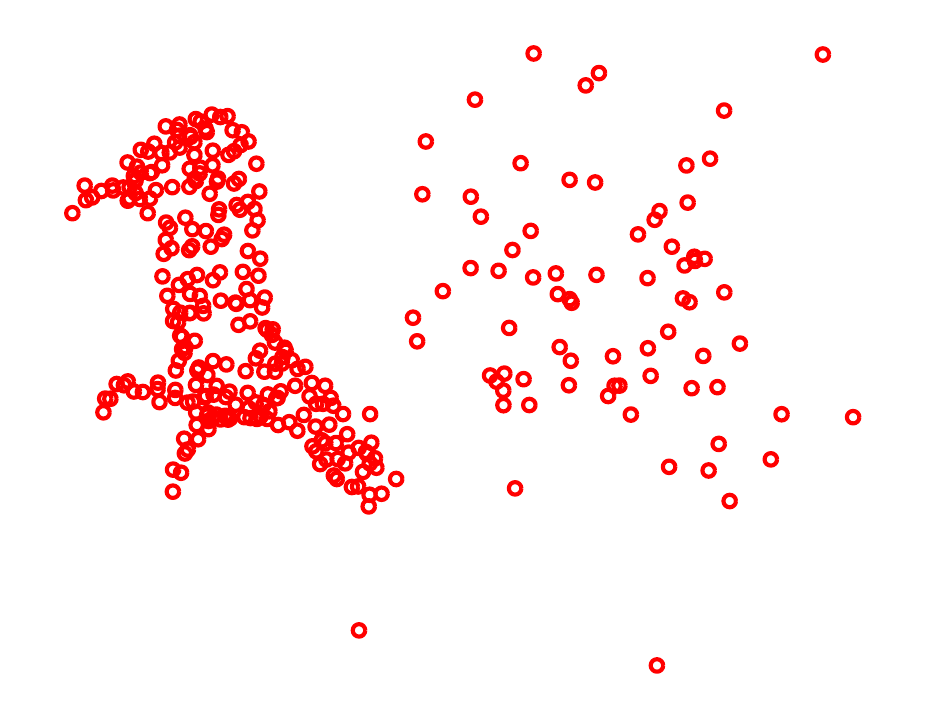}&
				\includegraphics[width=\scale\linewidth]{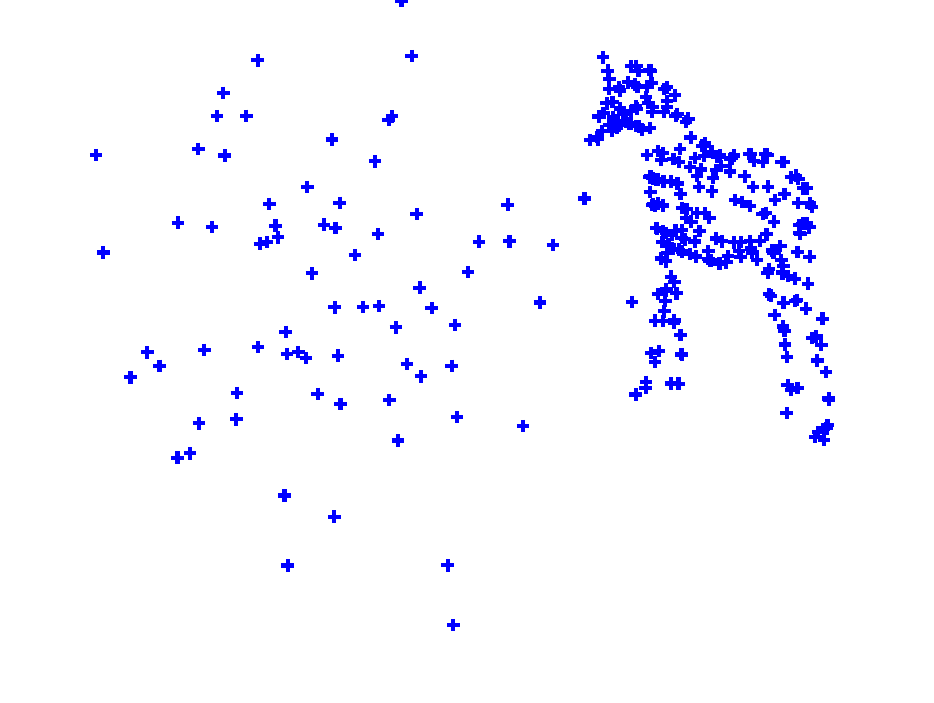}&
				\includegraphics[width=\scale\linewidth]{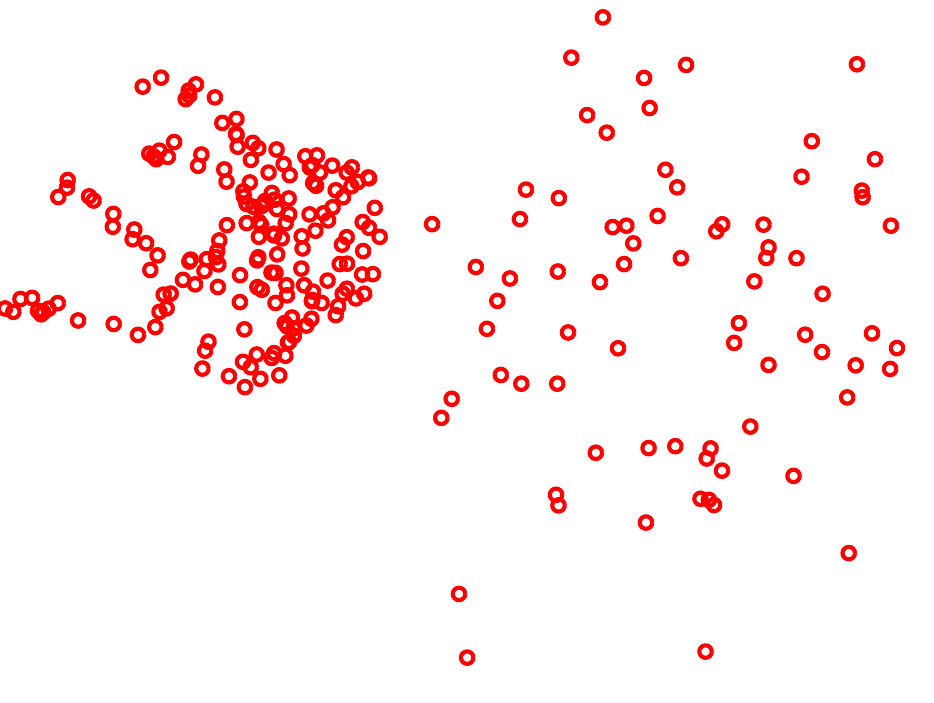}&
				\includegraphics[width=\scale\linewidth]{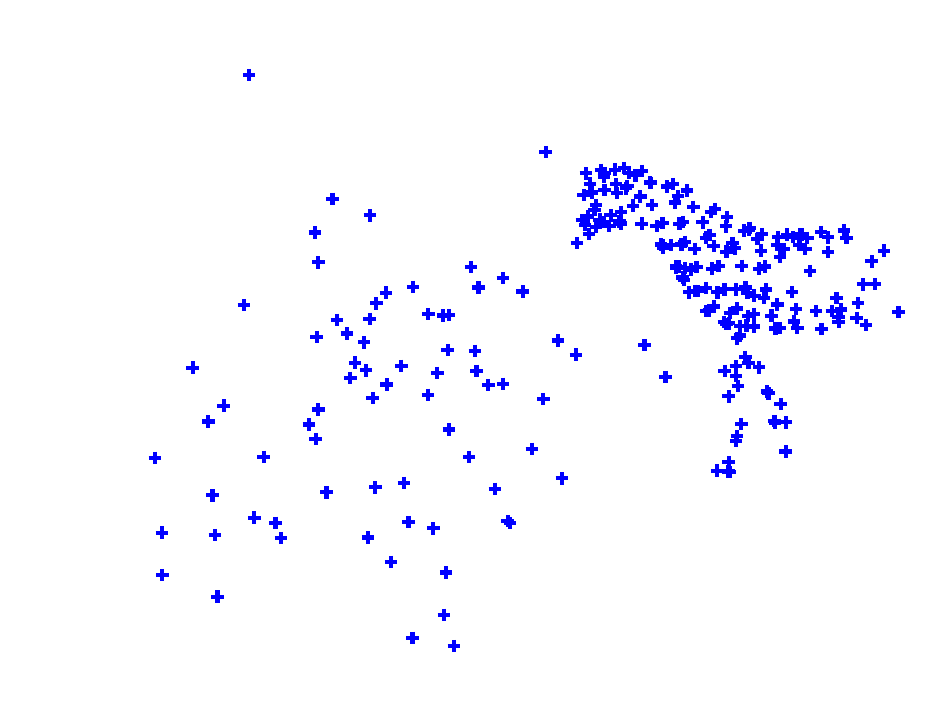} 
				\\\hline
				%
				{		
					\includegraphics[width=\scale\linewidth]{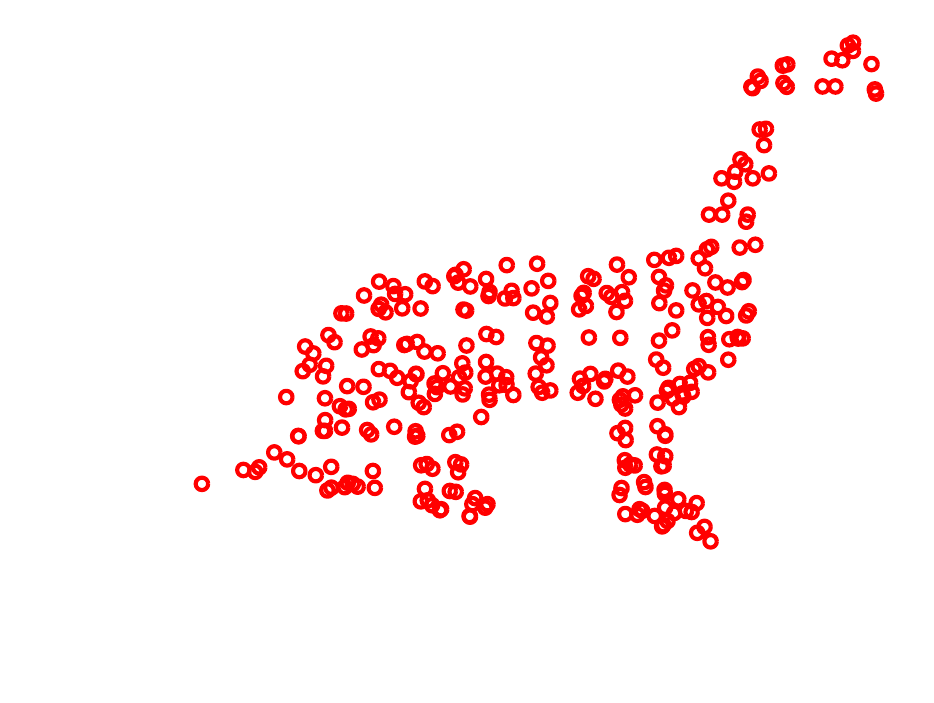}} &
				{		
					\includegraphics[width=\scale\linewidth]{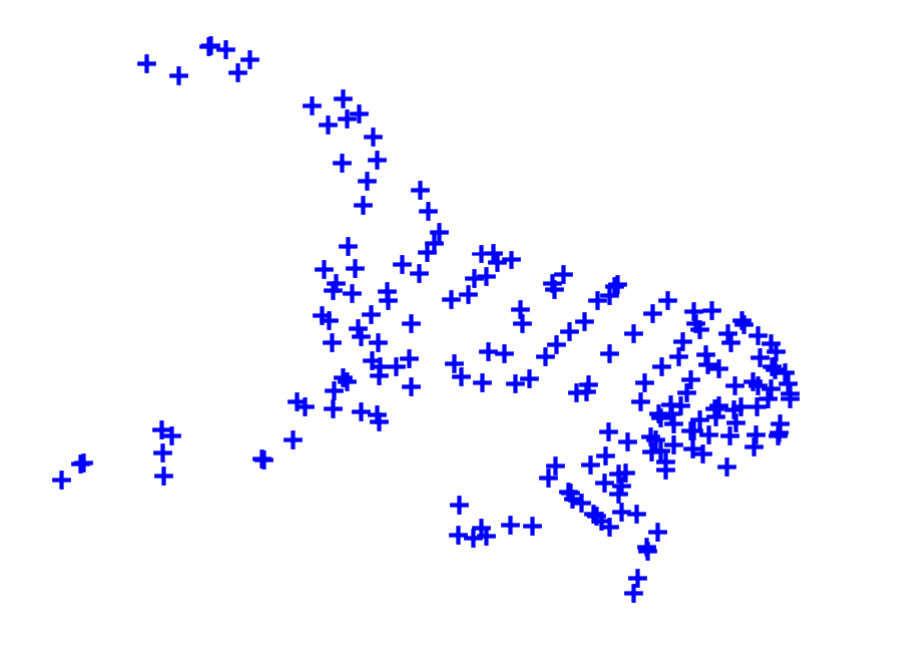}}&
				{		
					\includegraphics[width=\scale\linewidth]{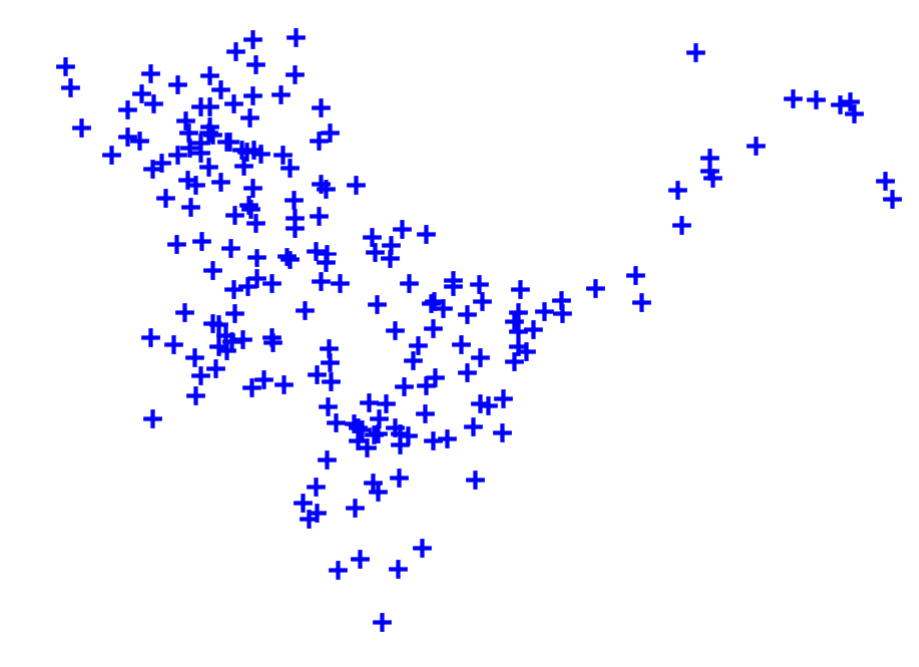}}&
				{\includegraphics[width=\scale\linewidth]{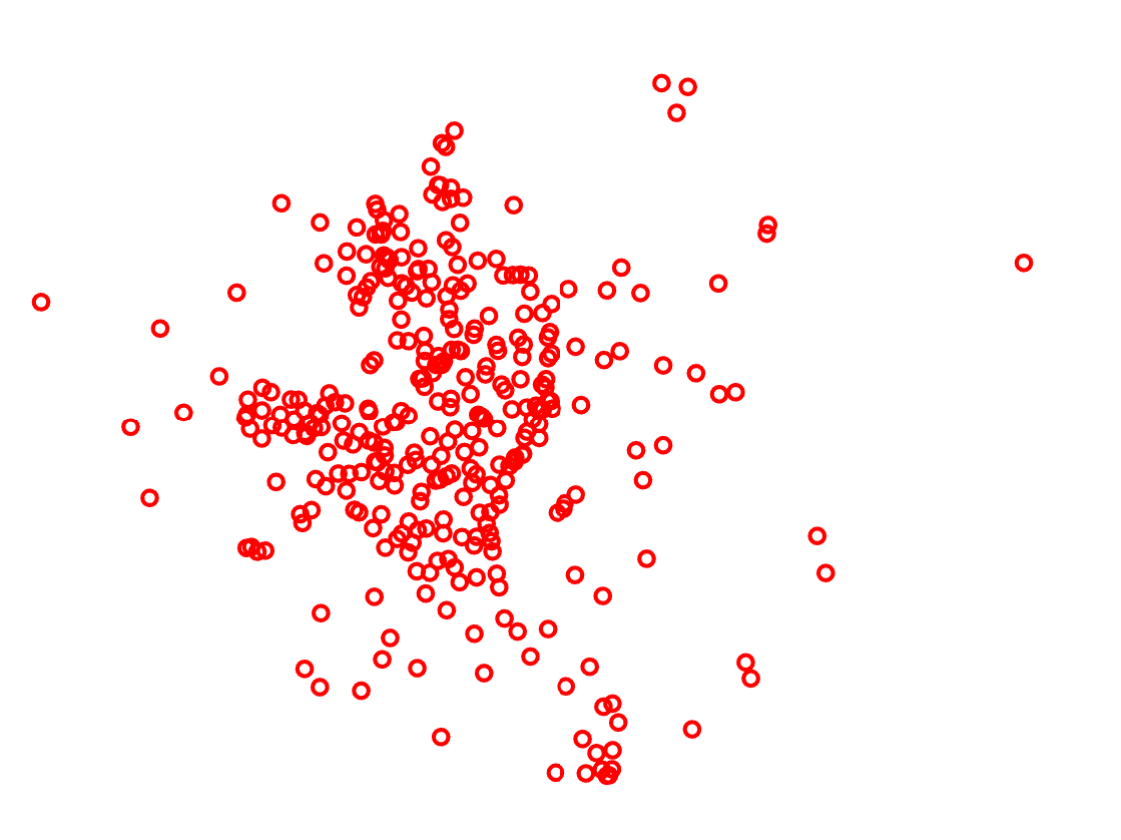}}&
				{\includegraphics[width=\scale\linewidth]{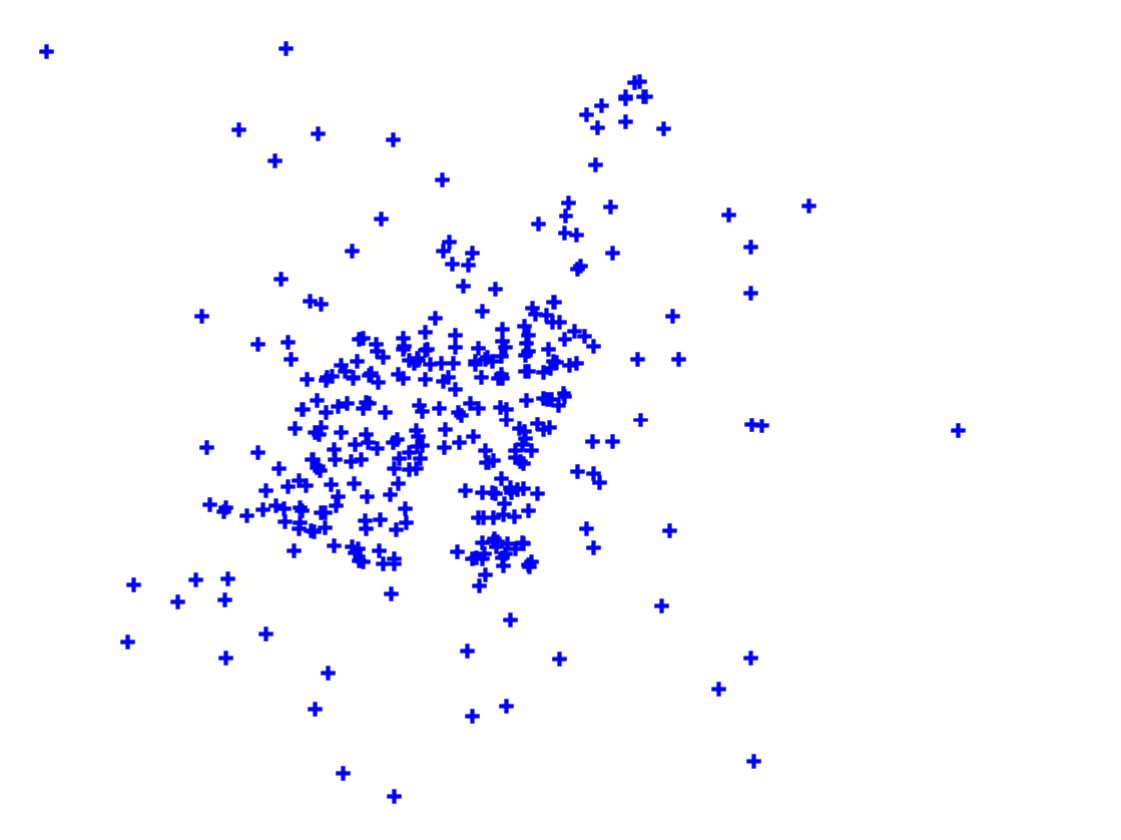}}&			
				{		
					\includegraphics[width=\scale\linewidth]{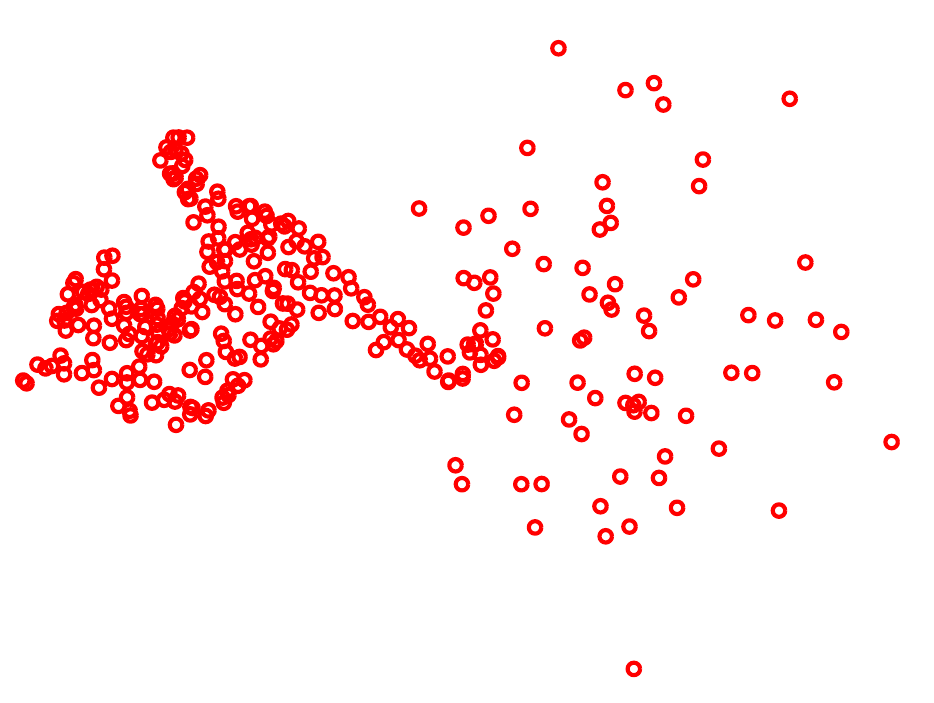}}&
				{		
					\includegraphics[width=\scale\linewidth]{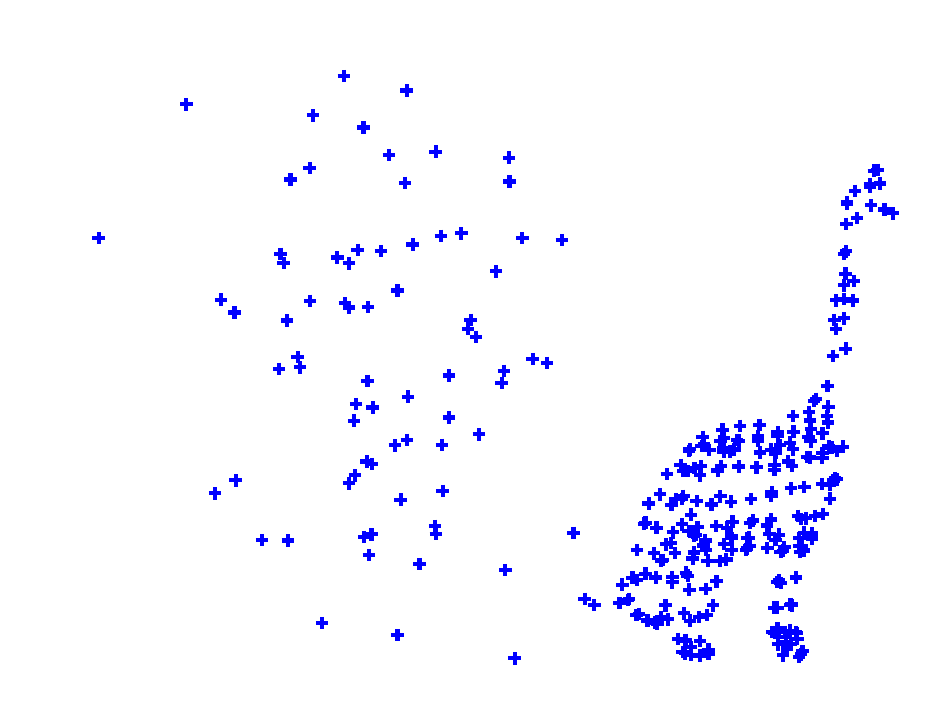}} &
				{		
					\includegraphics[width=\scale\linewidth]{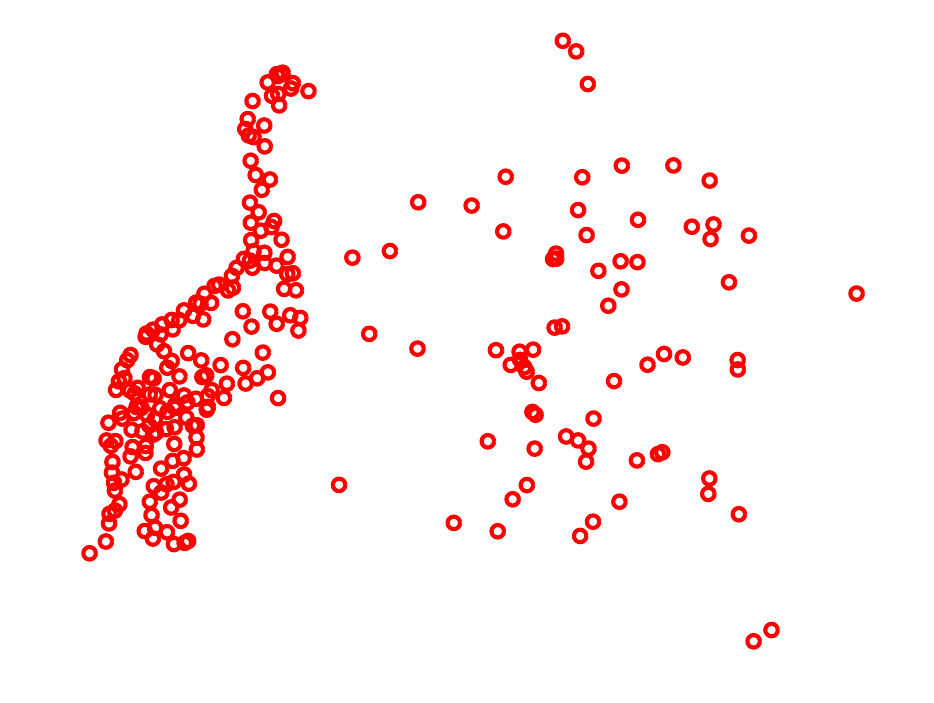}} &
				{
					\includegraphics[width=\scale\linewidth]{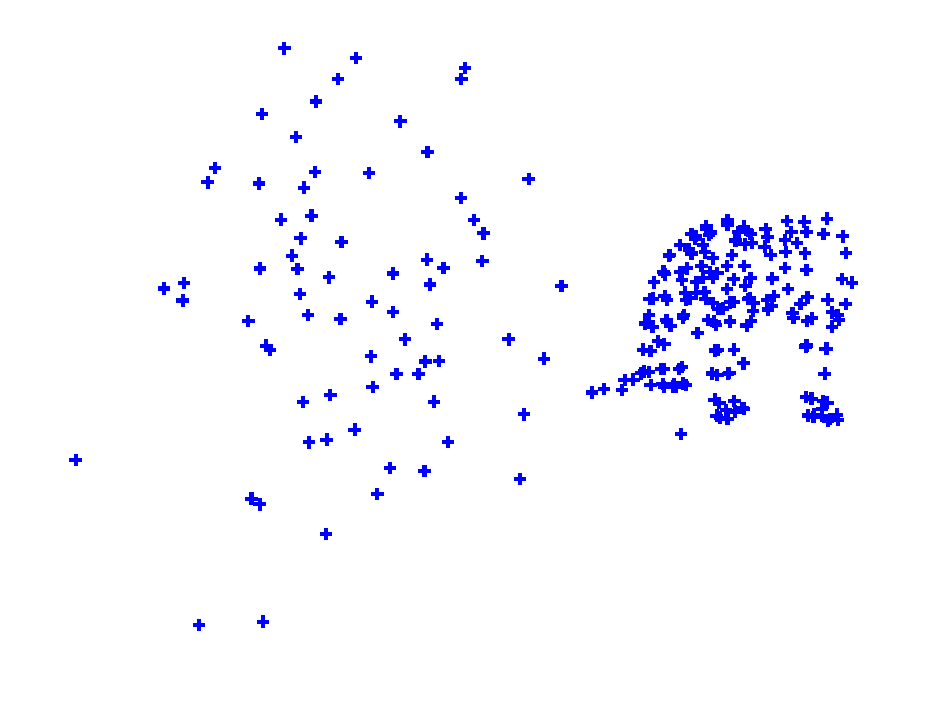}}
			\end{tabular}	
			\caption{
				(a) to (c): Source point sets and  examples of target point sets in the deformation and noise tests, respectively.
				(d) to (i): Examples of source and target point sets in the mixed outliers and inliers test ((d), (e)), separate outliers and inliers test ((f), (g)), and occlusion+outlier test ((h), (i)), respectively.
				In all cases, source points are indicated by red circles, while target points are represented by blue crosses.	
				\label{rot_3D_test_data_exa}}
			\centering
			\newcommand\scaleGd{10cm}
			\begin{tabular}{@{\hspace{-0mm}}c@{\hspace{-.7mm}}c@{\hspace{-.5mm}}c@{\hspace{-.6mm}}c@{\hspace{-1.2mm}} c }
				\includegraphics[height=\scaleGd]{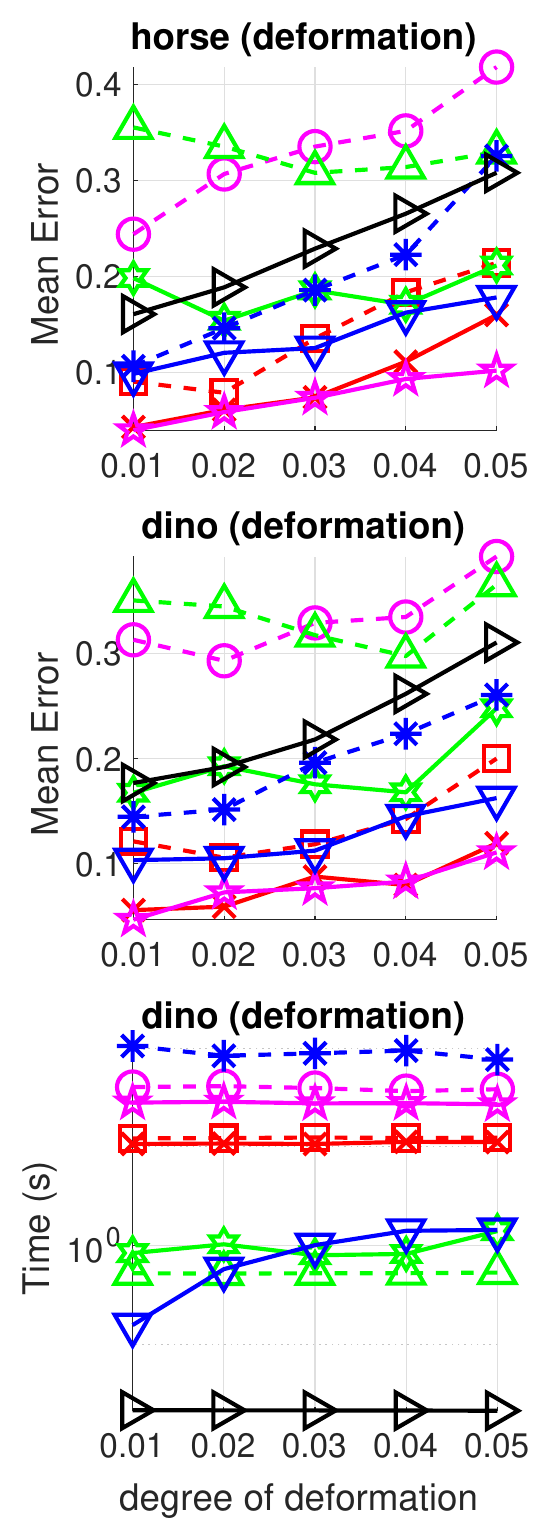}&	
				\includegraphics[height=\scaleGd]{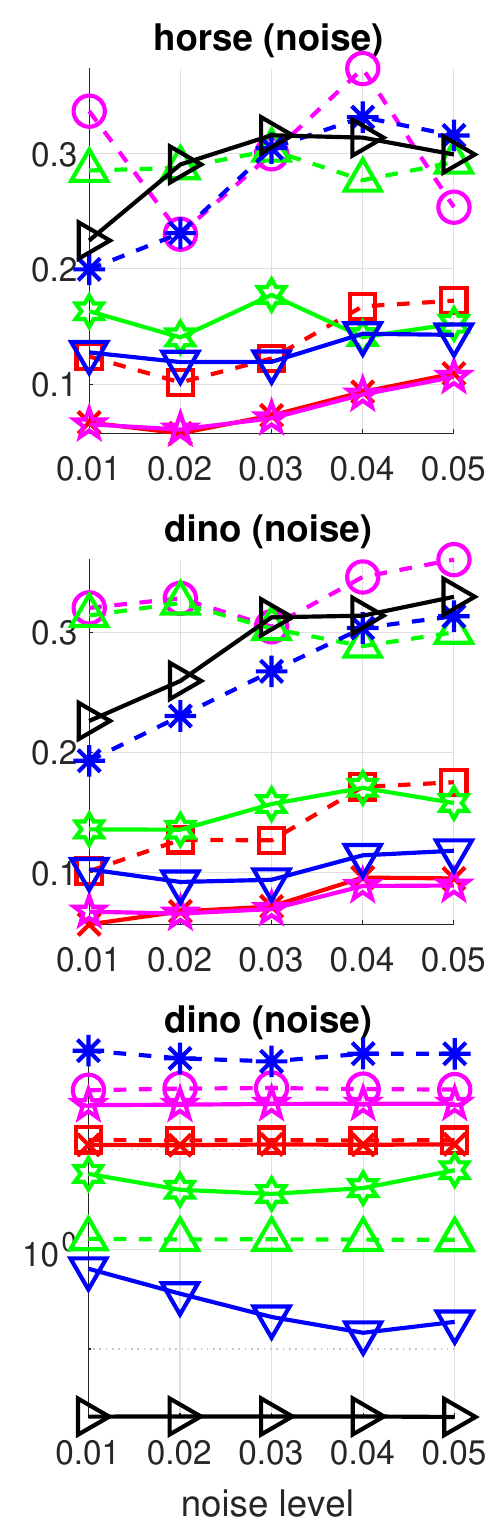}&		
				\includegraphics[height=\scaleGd]{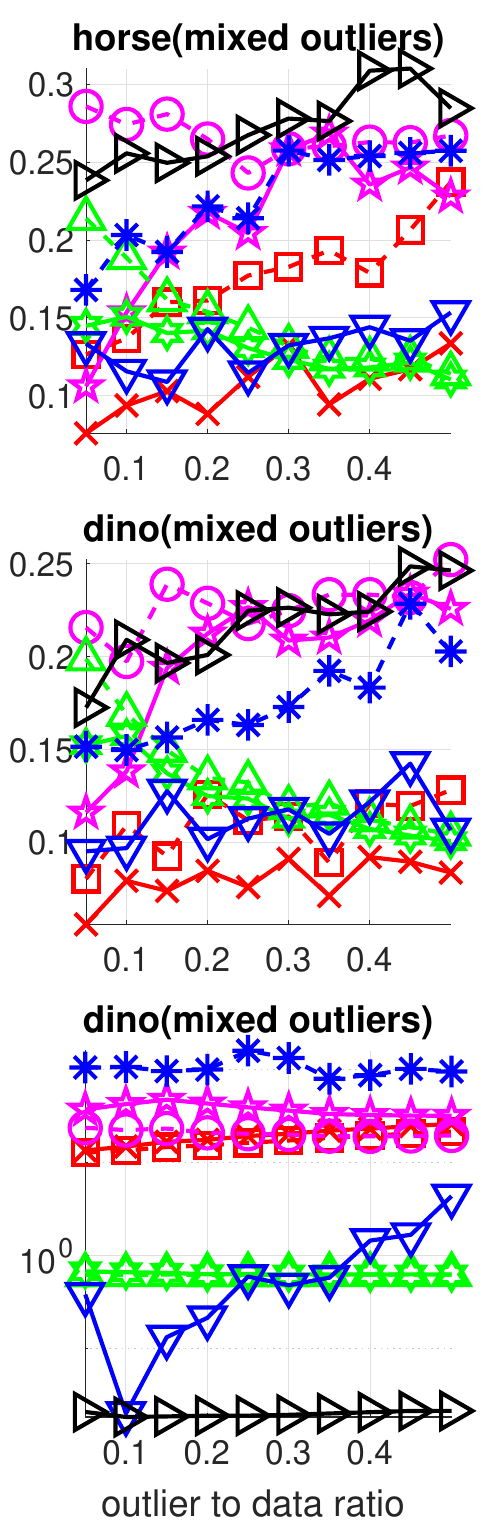}&
				
				\includegraphics[height=\scaleGd]{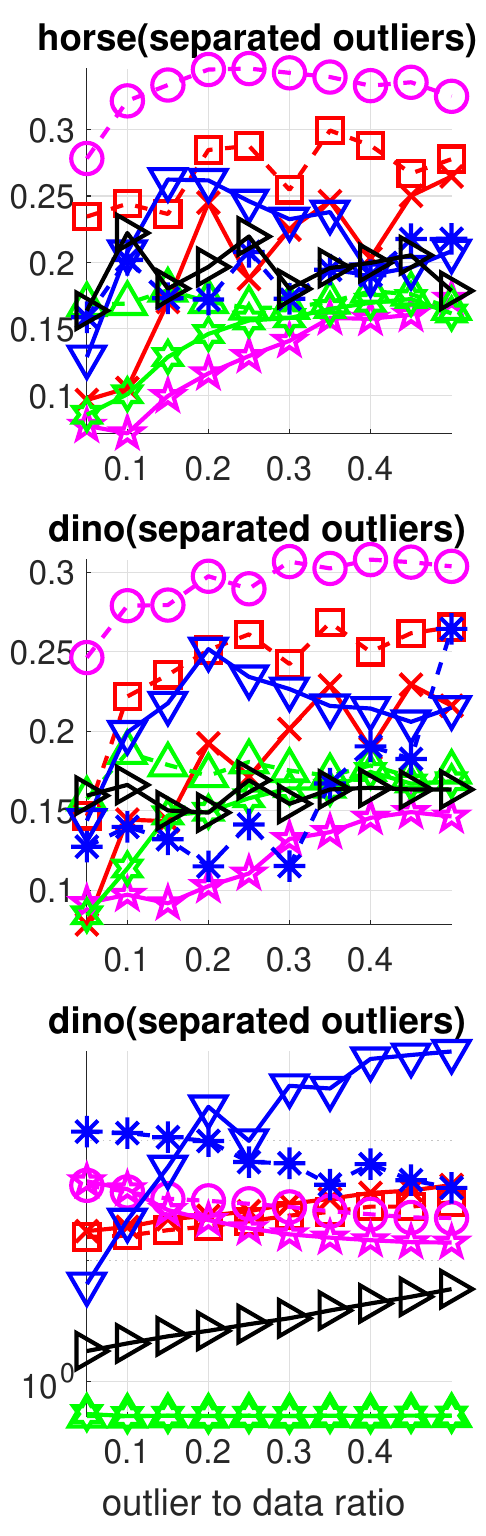}&
				\includegraphics[height=\scaleGd]{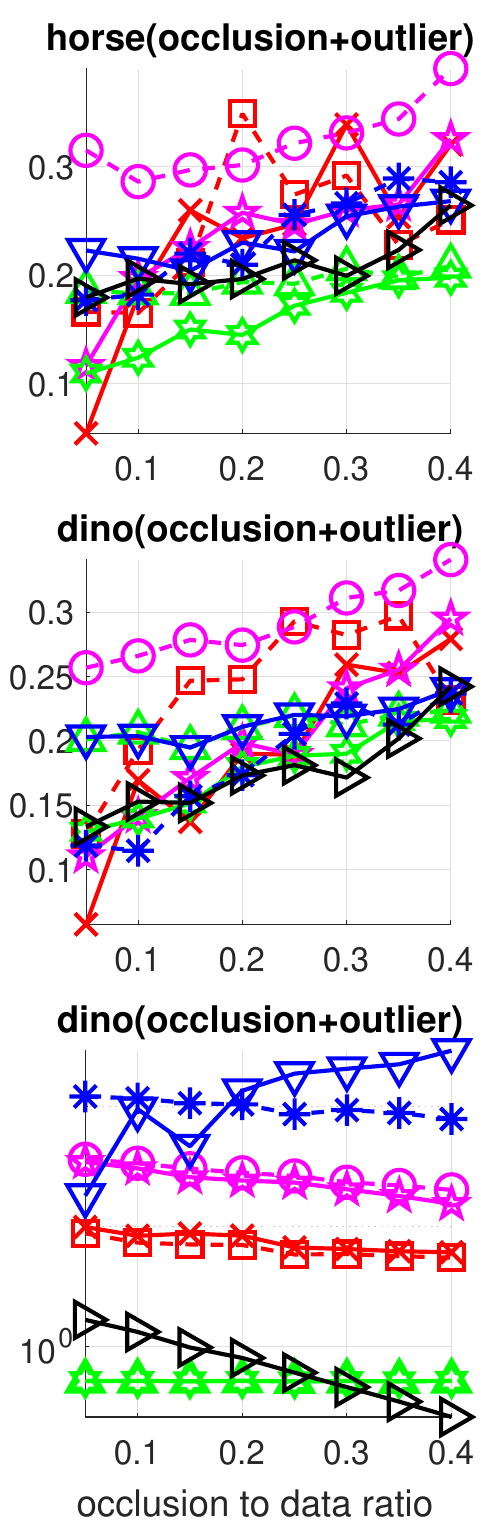}
				
			\end{tabular}
				\includegraphics[width=.7\linewidth]{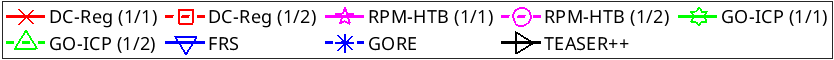}			
				\caption{
					Average registration errors (top 2 rows) and run times (bottom row) of
					various methods
					under varying $n_p$ values ($1/2$ or $1/1$ of the ground truth value) across 100 random trials for 3D deformation, positional noise, mixed outliers and inliers, separate outliers and inliers, and occlusion+outlier tests.			 
					\label{3D_rigid_sta}}
				\centering
				\includegraphics[width=1\linewidth]{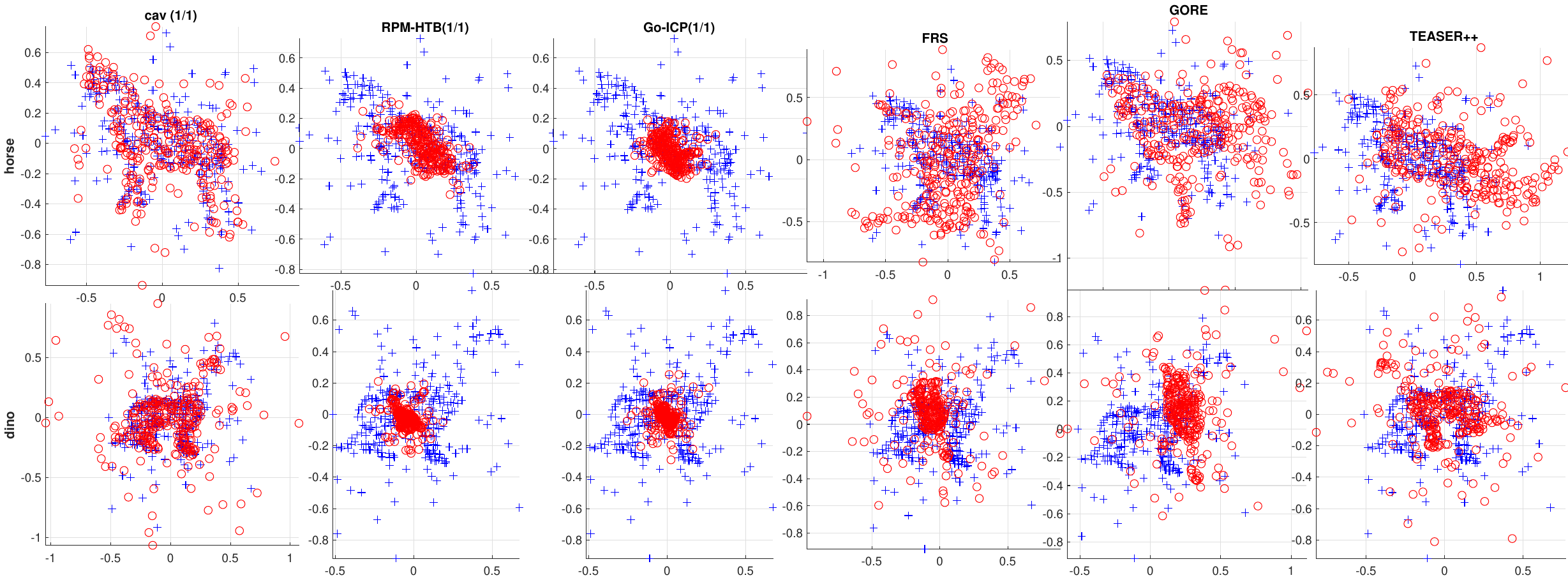}
				\caption{
					Examples of registration results from different methods in the separate outliers and inliers test, where the $n_p$ values of RPM-HTB and Go-ICP are both chosen as  the ground truth.		
					\label{rot_3D_syn_match_exa}}	
			\end{figure*}

		We compare DC-Reg ($n_p=1/1$ and $n_p=1/2$) against several state-of-the-art 3D registration methods, including RPM-HTB, GO-ICP, FRS, GORE, and TEASER++. Similar to the 2D experimental setup, we utilize "horse" and "dino" shapes to assess resilience to five disturbances: \textit{i)} Deformation, \textit{ii)} Noise, \textit{iii)} Mixed outliers, \textit{iv)} Separate outliers, and \textit{v)} Occlusion + Outlier tests. Fig. \ref{rot_3D_test_data_exa} provides a visual illustration of these tests, 
		Fig. \ref{3D_rigid_sta} presents statistical results,	
		while Fig. \ref{rot_3D_syn_match_exa} shows examples of registration results.

		\textbf{Resilience to Geometric Distortions.} As shown in the first two columns of Fig. \ref{3D_rigid_sta}, DC-Reg ($n_p=1/1$) (red solid cross) consistently yields the lowest registration errors across all levels of deformation and noise. In contrast, TEASER++ (black right-pointing triangle) and FRS (blue inverted triangle) exhibit increasing sensitivity and sharp error rises as the degree of disturbance intensifies. This performance degradation is primarily because these methods are strictly designed for rigid transformations; significant non-rigid deformations and high positional noise violate their underlying assumptions of distance-invariance, whereas DC-Reg’s holistic bound better accommodates such geometric distortions.

		\textbf{Outlier and Occlusion Handling.} The performance of DC-Reg varies depending on the outlier distribution:
		\begin{inparaenum}[\upshape i\upshape)]
			\item \textbf{Mixed Outliers:} DC-Reg ($n_p=1/1$) exhibits outstanding robustness in this scenario, maintaining the lowest error trajectory among all tested methods even as the outlier ratio reaches 0.5. This confirms that our global bound is highly effective at capturing global structure amidst uniformly distributed noise.
			\item \textbf{Separated Outliers:} For clustered outliers, DC-Reg shows moderate performance. While it remains significantly more stable than RPM-HTB, it is occasionally outperformed by GO-ICP or GORE in absolute precision. This suggests that concentrated outlier clusters pose a greater challenge to rotation-invariant feature (RIF) descriptors than scattered noise.
			\item \textbf{Occlusion + Outliers:} Under significant occlusion, DC-Reg remains stable but is not the top performer. Its error rises more sharply at high occlusion ratios compared to robust solvers like TEASER++ and GORE, which demonstrate superior resilience to large-scale missing data.	
		\end{inparaenum}

		\textbf{Parameter Sensitivity.} Despite an underestimated inlier count ($n_p=1/2$, red dashed square), DC-Reg demonstrates remarkable robustness. The $n_p=1/2$ variant remains significantly more accurate than RPM-HTB and achieves performance comparable to or better than GO-ICP. This stability validates that our holistic concave underestimator effectively captures the global spatial coupling, ensuring convergence even with poor cardinality estimates.

		\textbf{Computational Efficiency.} The bottom row of Fig. \ref{3D_rigid_sta} summarizes the average run times. While DC-Reg (red curves) is generally slower than TEASER++ and GO-ICP, it exhibits remarkable computational stability. Specifically, GORE (blue star) and RPM-HTB (magenta star) maintain consistently high computational costs, whereas DC-Reg operates at a lower time complexity in "Deformation", "Noise", and "Mixed Outliers" scenarios. Notably, in the "Separated Outliers" and "Occlusion + Outlier" tests, the run time of FRS (blue inverted triangle) escalates sharply as disturbances intensify, eventually exceeding that of DC-Reg in the later stages. This indicates that our method possesses superior resilience to extreme clutter. Furthermore, the efficiency of DC-Reg is nearly identical for both $n_p=1/1$ (solid cross) and $n_p=1/2$ (dashed square) variants, highlighting a consistent search complexity that is far less sensitive to inlier count priors than GO-ICP.

			\subsubsection{Evaluation on the 3DMatch Dataset}
			
\begin{figure*}[!ht]
					\centering
					\begin{tabular}{@{\hspace{-4mm}}c} 
						\includegraphics[width=1.08\linewidth]{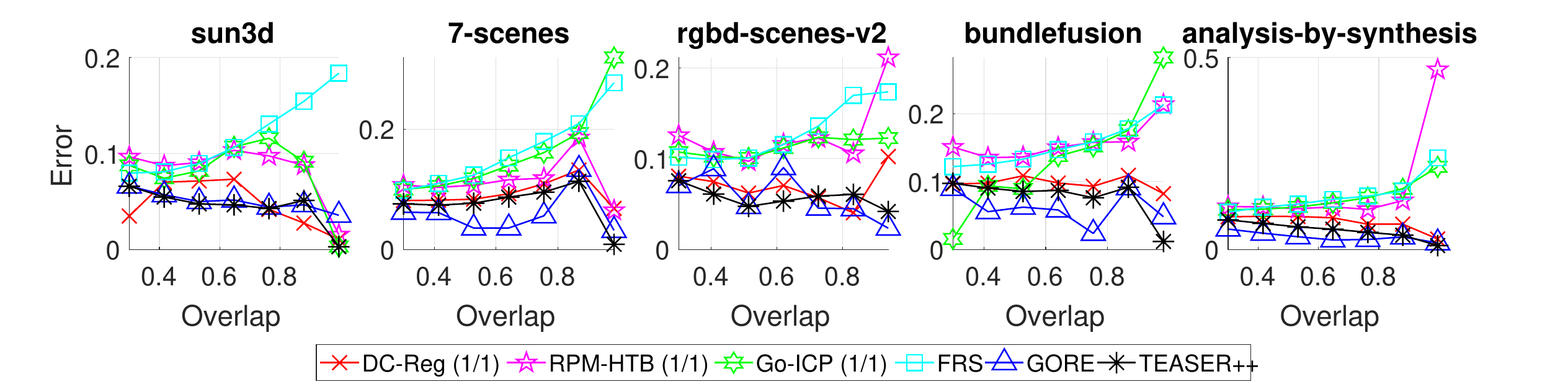}	
					\end{tabular}	
					\caption{
						Average registration errors from various methods across five RGB-D reconstruction datasets.
						\label{3DMatch_tests}	}
					\centering
					\newcommand\scale{1}	
					\includegraphics[width=\scale\linewidth]{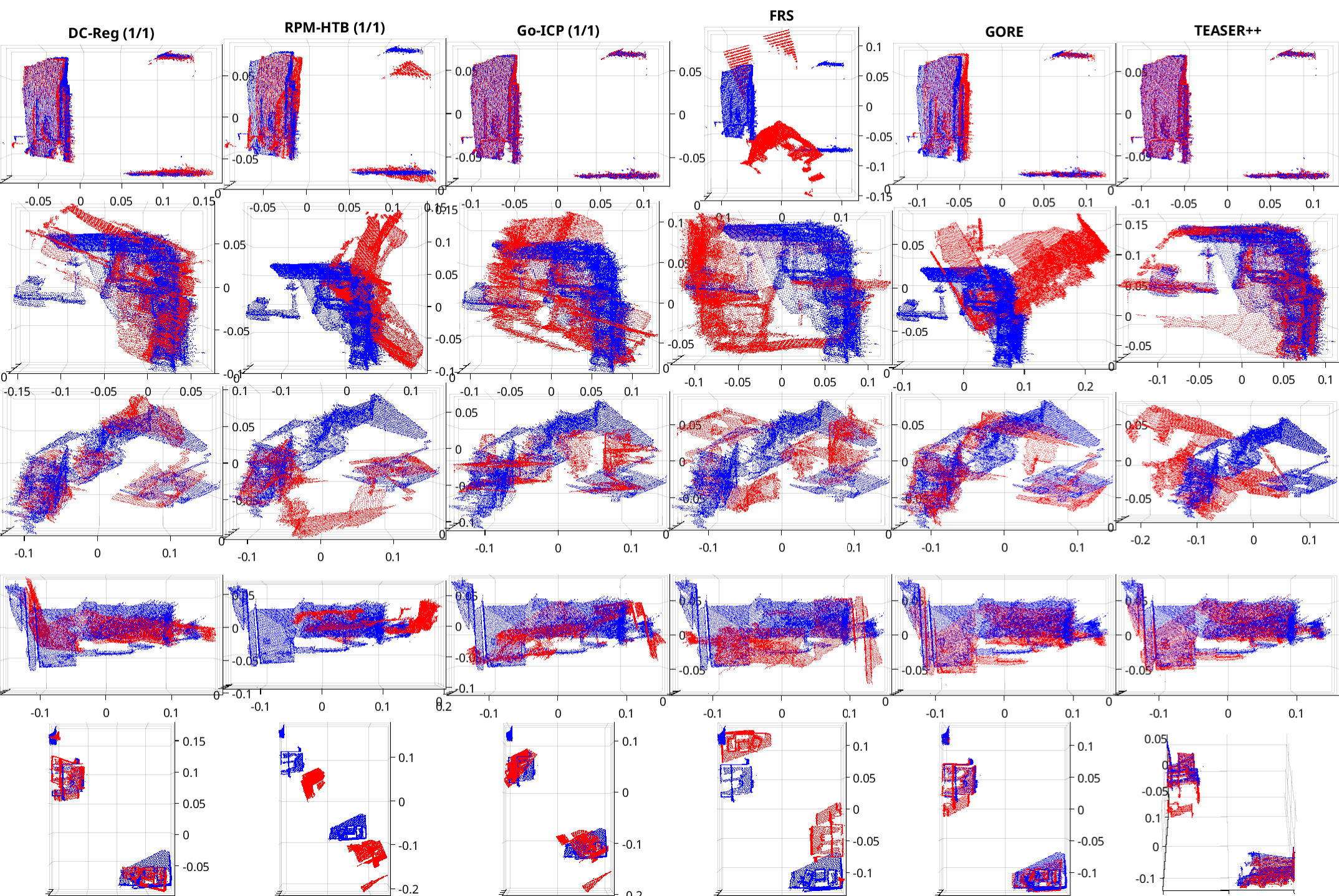}
					\caption{
						Examples of registration results generated by different methods on the five RGB-D reconstruction datasets, arranged from top to bottom: sun3d, 7-scenes, rgbd-scenes-v2 (repeated), and analysis-by-synthesis.
						\label{3DMatch_exa}	}
			\end{figure*}

			To evaluate the performance of our framework in real-world scenarios, we conduct experiments on the 3DMatch benchmark \cite{3Dmatch_dataset}, which comprises 62 scenes from five distinct RGB-D reconstruction datasets  (sun3d, 7-scenes, rgbd-scenes-v2, bundlefusion, and analysis-by-synthesis). 
			Following the standard evaluation protocol, 
			we utilize the specific scan pairs established by D3Feat \cite{D3feat_dataset} to 	  
			to evaluate the performance of different methods.
			The registration errors relative to different overlap ratios are illustrated in 	Fig. \ref{3DMatch_tests}.
			Examples of registration results are presented in Fig. \ref{3DMatch_exa}.
			
			As illustrated in Fig. 	\ref{3DMatch_tests},
			DC-Reg (red solid line) demonstrates exceptional robustness and consistently maintains a low registration error.	
			In the 7-scenes and analysis-by-synthesis datasets, DC-Reg maintains a nearly horizontal error trajectory near zero, indicating high stability regardless of the overlap ratio.
			Even in challenging low-overlap regimes (overlap $< 0.4$), DC-Reg effectively avoids local minima that trap other solvers like FRS or Go-ICP.
			
			
			\textbf{DC-Reg vs. RPM-HTB}: Our method significantly outperforms  RPM-HTB (magenta), across every test case. This confirms that our holistic concave underestimator provides a much tighter lower bound than the term-wise relaxations used in RPM-HTB, particularly when processing real-world scan noise.
			
			\textbf{DC-Reg vs. TEASER++ and GORE}: DC-Reg achieves performance parity with leading robust solvers like TEASER++ (black) and GORE (blue). In the sun3d dataset, DC-Reg exhibits a competitive downward error trend as overlap increases, similar to TEASER++.
			

	\section{Conclusion \label{sec:conclude}}

In this paper, we presented DC-Reg, a globally optimal framework for simultaneous pose and correspondence estimation. By reformulating the registration objective through the lens of Difference of Convex (DC) programming, we derived a holistic concave underestimator that captures the structural coupling between transformation and matching variables. This approach overcomes the fundamental limitations of previous Branch-and-Bound (BnB) solvers that rely on loose, term-wise relaxations.

Our extensive evaluations on 2D and 3D synthetic datasets, as well as real-world benchmarks, demonstrate that DC-Reg achieves superior convergence and robustness. Notably, unlike existing high-performance solvers strictly designed for rigid transformations, our method maintains exceptional precision under significant non-rigid deformations and high positional noise by effectively accommodating violations of distance-invariance.

Furthermore, experimental results confirm that DC-Reg provides not only a significantly tighter lower bound for global convergence but also remarkable computational stability. While the integration of Rotation-Invariant Features (RIF) involves a trade-off in absolute speed compared to decoupled solvers, DC-Reg’s execution time remains highly consistent even with inaccurate inlier count estimates ($n_p$), showcasing a predictable search complexity that is essential for reliable deployment in cluttered environments.

Future work will explore the extension of this DC programming paradigm to more complex non-rigid deformation models and the further acceleration of the BnB search using GPU-parallelized pruning strategies to mitigate the computational overhead in high-dimensional scenarios.

{\small
	\bibliography{DP_SC_CVPR}

@article{besl1992method,
	title={A method for registration of {3D} shapes},
	author={Besl, Paul J and McKay, Neil D},
	journal={IEEE Transactions on Pattern Analysis and Machine Intelligence},
	volume={14},
	number={2},
	pages={239--256},
	year={1992},
	publisher={IEEE}
}

@inproceedings{rusinkiewicz2001efficient,
	title={Efficient variants of the {ICP} algorithm},
	author={Rusinkiewicz, Szymon and Levoy, Marc},
	booktitle={Proceedings of the International Conference on {3-D} Digital Imaging and Modeling},
	pages={145--152},
	year={2001}
}

@inproceedings{aftab2015convergence,
	title={Convergence of iteratively re-weighted least squares to robust {M}-estimators},
	author={Aftab, Khurrum and Hartley, Richard},
	booktitle={Proceedings of the IEEE Winter Conference on Applications of Computer Vision},
	pages={313--320},
	year={2015}
}

@article{peng2023block,
	title={Block coordinate descent on smooth manifolds},
	author={Peng, Liangzu and Vidal, Ren{\'e}},
	journal={arXiv preprint arXiv:2305.14744},
	year={2023}
}

@inproceedings{peng2022global,
	title={Global linear and local superlinear convergence of {IRLS} for nonsmooth robust regression},
	author={Peng, Liangzu and K{\"u}mmerle, Christian and Vidal, Ren{\'e}},
	booktitle={Advances in Neural Information Processing Systems (NeurIPS)},
	pages={26262--26275},
	year={2022}
}

@article{yang2020graduated,
	title={Graduated non-convexity for robust spatial perception: {From} non-minimal solvers to global outlier rejection},
	author={Yang, Heng and Antonante, Pasquale and Tzoumas, Vasileios and Carlone, Luca},
	journal={IEEE Robotics and Automation Letters},
	volume={5},
	number={2},
	pages={1127--1134},
	year={2020}
}

@inproceedings{zhou2016fast,
	title={Fast global registration},
	author={Zhou, Qian-Yi and Park, Jaesik and Koltun, Vladlen},
	booktitle={Proceedings of the European Conference on Computer Vision (ECCV)},
	pages={766--782},
	year={2016}
}

@inproceedings{choy2020deep,
	title={Deep global registration},
	author={Choy, Christopher and Dong, Wei and Koltun, Vladlen},
	booktitle={Proceedings of the IEEE/CVF Conference on Computer Vision and Pattern Recognition (CVPR)},
	pages={2514--2523},
	year={2020}
}

@inproceedings{qin2023deep,
	title={Deep graph-based spatial consistency for robust non-rigid point cloud registration},
	author={Qin, Zheng and Yu, Hao and Wang, Changjian and Peng, Yuxing and Xu, Kai},
	booktitle={Proceedings of the IEEE/CVF Conference on Computer Vision and Pattern Recognition (CVPR)},
	pages={5394--5403},
	year={2023}
}

@inproceedings{wang2019deep,
	title={Deep closest point: Learning representations for point cloud registration},
	author={Wang, Yue and Solomon, Justin M},
	booktitle={Proceedings of the IEEE/CVF International Conference on Computer Vision (ICCV)},
	pages={3523--3532},
	year={2019}
}

@article{yao2023hunter,
	title={Hunter: Exploring high-order consistency for point cloud registration with severe outliers},
	author={Yao, Runzhao and Du, Shaoyi and Cui, Wenting and Ye, Aixue and Wen, Feng and Zhang, Hongbo and Tian, Zhiqiang and Gao, Yue},
	journal={IEEE Transactions on Pattern Analysis and Machine Intelligence},
	year={2023},
	publisher={IEEE}
}

@article{chen2023sc2pcrplus,
	title={{SC2-PCR++}: Rethinking the generation and selection for efficient and robust point cloud registration},
	author={Chen, Zhi and Sun, Kun and Yang, Fan and Guo, Lin and Tao, Wenbing},
	journal={IEEE Transactions on Pattern Analysis and Machine Intelligence},
	year={2023},
	publisher={IEEE}
}

@inproceedings{lusk2021clipper,
	title={{CLIPPER}: A graph-theoretic framework for robust data association},
	author={Lusk, Parker C and Fathian, Kaveh and How, Jonathan P},
	booktitle={Proceedings of the IEEE International Conference on Robotics and Automation (ICRA)},
	pages={13828--13834},
	year={2021}
}

@article{yang2021teaser,
	title={{TEASER}: Fast and certifiable point cloud registration},
	author={Yang, Heng and Shi, Jingnan and Carlone, Luca},
	journal={IEEE Transactions on Robotics},
	volume={37},
	number={2},
	pages={314--333},
	year={2021},
	publisher={IEEE}
}

@inproceedings{zhang20233d,
	title={{3D} registration with maximal cliques},
	author={Zhang, Xiyu and Yang, Jiaqi and Zhang, Shikun and Zhang, Yanning},
	booktitle={Proceedings of the IEEE/CVF Conference on Computer Vision and Pattern Recognition (CVPR)},
	pages={17745--17754},
	year={2023}
}

@article{parra2018guaranteed,
	title={Guaranteed outlier removal for point cloud registration with correspondences},
	author={Parra Bustos, {\'A}lvaro and Chin, Tat-Jun},
	journal={IEEE Transactions on Pattern Analysis and Machine Intelligence},
	volume={40},
	number={12},
	pages={2868--2882},
	year={2018},
	publisher={IEEE}
}

@inproceedings{chin2016guaranteed,
	title={Guaranteed outlier removal with mixed integer linear programs},
	author={Chin, Tat-Jun and Kee, Yang Heng and Eriksson, Anders and Neumann, Frank},
	booktitle={Proceedings of the IEEE Conference on Computer Vision and Pattern Recognition (CVPR)},
	pages={4458--4466},
	year={2016}
}

@inproceedings{briales2017convex,
	title={Convex global {3D} registration with {Lagrangian} duality},
	author={Briales, Jesus and Gonzalez-Jimenez, Javier},
	booktitle={Proceedings of the IEEE Conference on Computer Vision and Pattern Recognition (CVPR)},
	pages={4960--4969},
	year={2017}
}

@inproceedings{iglesias2020global,
	title={Global optimality for point set registration using semidefinite programming},
	author={Iglesias, Jose Pedro and Olsson, Carl and Kahl, Fredrik},
	booktitle={Proceedings of the IEEE/CVF Conference on Computer Vision and Pattern Recognition (CVPR)},
	pages={8287--8295},
	year={2020}
}

@inproceedings{peng2022semidefinite,
	title={Semidefinite relaxations of truncated least-squares in robust rotation search: Tight or not},
	author={Peng, Liangzu and Fazlyab, Mahyar and Vidal, Ren{\'e}},
	booktitle={European Conference on Computer Vision (ECCV)},
	pages={135--151},
	year={2022},
	organization={Springer}
}

@inproceedings{yang2019quaternion,
	title={A quaternion-based certifiably optimal solution to the {Wahba} problem with outliers},
	author={Yang, Heng and Carlone, Luca},
	booktitle={Proceedings of the IEEE/CVF International Conference on Computer Vision (ICCV)},
	pages={3105--3114},
	year={2019}
}

@book{scholz2011deterministic,
	title={Deterministic Global Optimization: Geometric Branch-and-bound Methods and Their Applications},
	author={Scholz, Daniel},
	year={2011},
	publisher={Springer Science \& Business Media},
	isbn={9781461419327}
}

@inproceedings{chen2022deterministic,
	title={Deterministic point cloud registration via novel transformation decomposition},
	author={Chen, Wen and Li, Haoang and Nie, Qiang and Liu, Yun-Hui},
	booktitle={Proceedings of the IEEE/CVF Conference on Computer Vision and Pattern Recognition},
	pages={17343--17351},
	year={2022}
}

@inproceedings{li20073d,
	title={The {3D-3D} registration problem revisited},
	author={Li, Hongdong and Hartley, Richard},
	booktitle={Proceedings of the IEEE International Conference on Computer Vision},
	pages={1--8},
	year={2007}
}

@inproceedings{liu2018efficient,
	title={Efficient global point cloud registration by matching rotation invariant features through translation search},
	author={Liu, Yinlong and Wang, Chen and Song, Zhijian and Wang, Manning},
	booktitle={European Conference on Computer Vision},
	pages={445--461},
	year={2018}
}

@article{olsson2008branch,
	title={Branch-and-bound methods for {Euclidean} registration problems},
	author={Olsson, Carl and Kahl, Fredrik and Oskarsson, Magnus},
	journal={IEEE Transactions on Pattern Analysis and Machine Intelligence},
	volume={31},
	number={5},
	pages={783--794},
	year={2008},
	publisher={IEEE}
}

@article{parra2016fast,
	title={Fast rotation search with stereographic projections for {3D} registration},
	author={Parra Bustos, {\'A}lvaro and Chin, Tat-Jun and Eriksson, Anders and Li, Hongdong and Suter, David},
	journal={IEEE Transactions on Pattern Analysis and Machine Intelligence},
	volume={38},
	number={11},
	pages={2227--2240},
	year={2016},
	publisher={IEEE}
}

@article{wang2022efficient,
	title={Efficient and outlier-robust simultaneous pose and correspondence determination by branch-and-bound and transformation decomposition},
	author={Wang, Chen and Liu, Yinlong and Wang, Yiru and Li, Xuechen and Wang, Manning},
	journal={IEEE Transactions on Pattern Analysis and Machine Intelligence},
	volume={44},
	number={10},
	pages={6924--6938},
	year={2022},
	publisher={IEEE}
}

@article{yang2016go,
	title={{Go-ICP}: A globally optimal solution to {3D ICP} point-set registration},
	author={Yang, Jiaolong and Li, Hongdong and Campbell, Dylan and Jia, Yunde},
	journal={IEEE Transactions on Pattern Analysis and Machine Intelligence},
	volume={38},
	number={11},
	pages={2241--2254},
	year={2016},
	publisher={IEEE}
}

@InProceedings{10.1007/978-3-030-01258-8_28,
	author="Liu, Yinlong
	and Wang, Chen
	and Song, Zhijian
	and Wang, Manning",
	editor="Ferrari, Vittorio
	and Hebert, Martial
	and Sminchisescu, Cristian
	and Weiss, Yair",
	title="Efficient Global Point Cloud Registration by Matching Rotation Invariant Features Through Translation Search",
	booktitle="Computer Vision -- ECCV 2018",
	year="2018",
	publisher="Springer International Publishing",
	address="Cham",
	pages="460--474",
	isbn="978-3-030-01258-8"
}

@article{Lian2024,
	author    = {Wei Lian and Fei Ma and Zhesen Cui and Hang Pan},
	title     = {{HBSP}: a hybrid bilinear and semidefinite programming approach for aligning partially overlapping point clouds},
	journal   = {Scientific Reports},
	year      = {2024},
	volume    = {14},
	number    = {1},
	pages     = {30044},
	issn      = {2045-2322},
	doi       = {10.1038/s41598-024-79744-x},
	url       = {https://doi.org/10.1038/s41598-024-79744-x}
}

@article{10.1007/s11263-020-01359-2,
	author = {Ma, Jiayi and Jiang, Xingyu and Fan, Aoxiang and Jiang, Junjun and Yan, Junchi},
	title = {Image Matching from Handcrafted to Deep Features: A Survey},
	year = {2021},
	issue_date = {Jan 2021},
	publisher = {Kluwer Academic Publishers},
	address = {USA},
	volume = {129},
	number = {1},
	issn = {0920-5691},
	url = {https://doi.org/10.1007/s11263-020-01359-2},
	doi = {10.1007/s11263-020-01359-2},
	journal = {Int. J. Comput. Vision},
	month = {jan},
	pages = {23-79},
	numpages = {57}
}

@article{D3feat_dataset,
	title={D3Feat: Joint Learning of Dense Detection and Description of 3D Local Features},
	author={Xuyang Bai and Zixin Luo and Lei Zhou and Hongbo Fu and Long Quan and Chiew-Lan Tai},
	journal={arXiv:2003.03164 [cs.CV]},
	year={2020}
}

@inproceedings{3Dmatch_dataset,
	title={3DMatch: Learning Local Geometric Descriptors from RGB-D Reconstructions},
	author={Zeng, Andy and Song, Shuran and Nie{\ss}ner, Matthias and Fisher, Matthew and Xiao, Jianxiong and
	Funkhouser, Thomas},
	booktitle={CVPR},
	year={2017}
}

@article{LIAN2023126482,
	title = {Hybrid trilinear and bilinear programming for aligning partially overlapping point sets},
	journal = {Neurocomputing},
	volume = {551},
	pages = {126482},
	year = {2023},
	issn = {0925-2312},
	doi = {https://doi.org/10.1016/j.neucom.2023.126482},
	url = {https://www.sciencedirect.com/science/article/pii/S0925231223006057},
	author = {Wei Lian and Wangmeng Zuo}
}

@ARTICLE{lian2021polyhedral,
author={Lian, Wei and Zuo, Wangmeng and Cui, Zhesen},
journal={IEEE Access}, 
title={A Polyhedral Annexation Algorithm for Aligning Partially Overlapping Point Sets}, 
year={2021},
volume={9},
number={},
pages={166750-166761},
doi={10.1109/ACCESS.2021.3135863}
}

@article{RPM_model_occlude_PR,
title = {A concave optimization algorithm for matching partially overlapping point sets},
journal = {Pattern Recognition},
volume = {103},
pages = {107322},
year = {2020},
issn = {0031-3203},
doi = {https://doi.org/10.1016/j.patcog.2020.107322},
author = {Wei Lian and Lei Zhang},
}

@misc{caltech_database,
author={G. Griffin and A. Holub and P. Perona},
title={Caltech-256 Object Category Dataset},
note={technical report, California Inst. of Technology},
year={2007}
}

@misc{pascal-voc-2007,
author ={Everingham, M. and Van~Gool, L. and Williams, C. K. I. and Winn, J. and Zisserman, A.},
title ={The {PASCAL} {V}isual {O}bject {C}lasses {C}hallenge 2007 {(VOC2007)} {R}esults},
howpublished ={http://www.pascal-network.org/challenges/VOC/voc2007/workshop/index.html}
}

@ARTICLE{Go-ICP_pami,
	
	author={Yang, Jiaolong and Li, Hongdong and Campbell, Dylan and Jia, Yunde},
	
	journal={IEEE Transactions on Pattern Analysis and Machine Intelligence}, 
	
	title={Go-ICP: A Globally Optimal Solution to 3D ICP Point-Set Registration}, 
	
	year={2016},
	
	volume={38},
	
	number={11},
	
	pages={2241-2254},
	
	doi={10.1109/TPAMI.2015.2513405}}

@inproceedings{Go-ICP,
author={Jiaolong Yang and Hongdong Li and  Yunde Jia},
title={Go-ICP: Solving 3D Registration Efficiently and Globally Optimally},
booktitle={International Conference on Computer Vision},
year={2013}
}

@article{CPD_match,
title={Point Set Registration: Coherent Point Drift},
author={Andriy Myronenko and Xubo Song},
journal={IEEE Transactions on Pattern Analysis and Machine Intelligence},
volume=32,
number=12,
year=2010,
pages={2262-2275}
}

@article{ICP,
journal={IEEE Trans. Pattern Analysis and Machine Intelligence},
 author = {Besl, Paul J. and McKay, Neil D.},
 title = {A Method for Registration of 3-D Shapes},
 volume = {14},
 number = {2},
 year = {1992},
 pages = {239--256},
}

@article{RPM_TPS,
journal={ Computer Vision and Image Understanding}, 
volume= 89,
pages={114-141},
year= 2003,
 author = {Chui, Haili and Rangarajan, Anand},
 title = {A new point matching algorithm for non-rigid registration},
 number = {2-3},
}

@inproceedings{RPM_concave,
author={Wei Lian and Lei Zhang},
title={Robust point matching revisited: a concave optimization approach},
booktitle={European conference on computer vision}, 
pages={ }, 
year=2012
}

@ARTICLE{RPM_concave_PAMI,
	author={Lian, Wei and Zhang, Lei and Yang, Ming-Hsuan},
	journal={IEEE Transactions on Pattern Analysis and Machine Intelligence}, 
	title={An Efficient Globally Optimal Algorithm for Asymmetric Point Matching}, 
	year={2017},
	volume={39},
	number={7},
	pages={1281-1293}
	}

@inproceedings{RPM_model_occlude,
author={Wei Lian and Lei Zhang},
booktitle={IEEE Conference on Computer Vision and Pattern Recognition},
title={Point Matching in the Presence of Outliers in Both Point Sets: A Concave Optimization Approach},
year={2014},
volume={},
number={},
pages={352-359}, 
}
}

\end{document}